\documentclass{article}
\usepackage{ijcai25}

\usepackage{times}
\usepackage{soul}
\usepackage{url}
\usepackage[hidelinks]{hyperref}
\usepackage[utf8]{inputenc}
\usepackage[small]{caption}
\usepackage{graphicx}
\usepackage{amsmath}
\usepackage{amsthm}
\usepackage{booktabs}
\usepackage{algorithm}
\usepackage{algorithmic}
\usepackage[switch]{lineno}

\usepackage{amssymb}
\usepackage{subfig}
\usepackage{setspace}
\usepackage{amsmath}

\usepackage{multirow}

\title{One Head Eight Arms: Block Matrix based Low Rank Adaptation for CLIP-based Few-Shot Learning}

\author{
Chunpeng Zhou$^1$
\and
Qianqian Shen$^1$\and
Zhi Yu$^1$\and
Jiajun Bu $^1$ \And
Haishuai Wang $^1$\\
\affiliations
$^1$Zhejiang university\\
\emails
\{zhoucp, shenqq377, yuzhirenzhe, bjj\}@zju.edu.cn,
haishuai.wang@gmail.com
}

\begin{document}

\maketitle

\begin{abstract}
Recent advancements in fine-tuning Vision-Language Foundation Models (VLMs) have garnered significant attention for their effectiveness in downstream few-shot learning tasks.While these recent approaches exhibits some performance improvements, they often  suffer from excessive training parameters and high computational costs.  To address these challenges, we propose a novel Block matrix-based low-rank adaptation framework, called Block-LoRA, for fine-tuning VLMs on downstream few-shot tasks. Inspired by recent work on Low-Rank Adaptation (LoRA), Block-LoRA partitions the original low-rank decomposition matrix of LoRA into a series of sub-matrices while sharing all down-projection sub-matrices. This structure not only reduces the number of training parameters, but also transforms certain complex matrix multiplication operations into simpler matrix addition, significantly lowering the computational cost of fine-tuning. 
Notably, Block-LoRA enables fine-tuning CLIP on the ImageNet few-shot benchmark using a single 24GB GPU. 
We also show that Block-LoRA has the more
tighter bound of generalization error than vanilla LoRA.
Without bells and whistles, extensive experiments demonstrate that Block-LoRA achieves competitive performance compared to state-of-the-art CLIP-based few-shot methods, while maintaining a low training parameters count and reduced computational overhead. \footnote{Under Review}
\end{abstract}

\section{Introduction}
In recent years, research on Foundation Models has made significant progress, demonstrating strong generalization capabilities and remarkable performance across various downstream tasks \cite{bommasani2021opportunities,moor2023foundation,zhou2024comprehensive,gu2023systematic,zhang2024vision,awais2023foundational,long2022vision}. In particular, Vision-Language Foundation Models (VLMs) have shown impressive performance in Computer Vision tasks, such as CLIP \cite{radford2021learning}, ALIGN \cite{jia2021scaling}.
\begin{figure}[t]
	\centering
	\includegraphics[width=1.0\columnwidth]{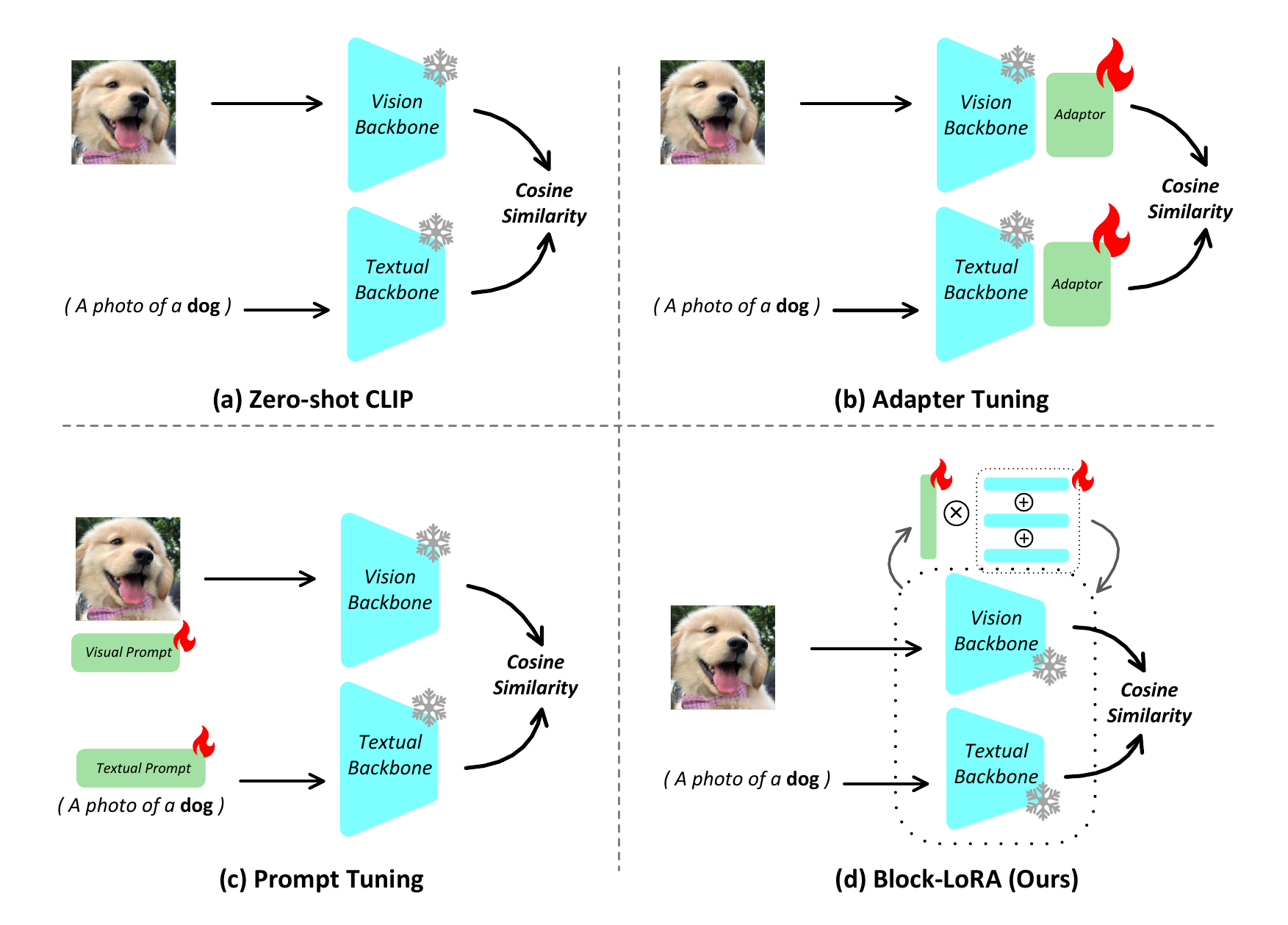}
	\caption{Architecture comparison of the different CLIP-based few-shot learning methods with ours Block-LoRA.}
	\label{fig:1}
\vspace{-0.5cm}
\end{figure}
Consequently, recent studies have explored leveraging VLMs for few-shot learning. Given the substantial number of parameters in VLMs, most approaches  adopt  the Parameter Efficient Fine-Tuning (PEFT) techniques \cite{han2024parameter} to fine-tune VLMs to to adapt these models to downstream few-shot tasks.
One prominent category of PEFT methods is adapter-based fine-tuning \cite{Clip-adapter,Tip-adapter,Meta-adapter,gondal2024domain}, which integrates learnable adapter modules into CLIP to enhance its performance on few-shot tasks, as shown in Fig \ref{fig:1}(b). For example, Meta-Adapter \cite{Meta-adapter} employs the self-attention based adapter modules  to refine features.
Another line focuses on the prompt-based fine-tuning \cite{zhou2022learning,zhou2022conditional,khattak2023maple,khattak2023self} to improve the few-shot performance by introducing the additional learnable vectors, , as shown in Fig \ref{fig:1}(c). 
For example, MaPle \cite{khattak2023maple} and
 PromptSRC \cite{khattak2023self} leverage  multi-modal Prompt Learning to enhance both vision and language features.  
While these adapter-based and prompt-based fine-tuning improve the performance of CLIP-based few-shot classification tasks, they introduce  excessive computational overhead and increase inference latency due to the extra computing modules incorporated into the CLIP model.
The recent CLIP-LoRA \cite{zanella2024low} directly employs Low-Rank Adaptation (LoRA) modules \cite{hu2022lora} to fine-tune CLIP for few-shot learning.
The deployment of LoRA has demonstrated substantial
improvements while significantly reducing training overhead, making it a promising approach for efficient few-shot adaptation.  
However, recent studies on LoRA has revealed the redundancy in LoRA's structure \cite{zhu2024asymmetry,zhang2023lora}. 
Experimental results on certain textual  tasks indicate that the dimensionality reduction matrix (usually denoted as $\mathbf{A}$) is typically task-irrelevant. Even when $\mathbf{A}$ remains randomly initialized and untrained, some downstream tasks remains close to that of the vanilla LoRA \cite{zhang2023lora}.

Building upon these insights,  we propose a novel Block Matrix based low rank adaptation framework, called Block-LoRA, designed to reduce redundancy, for few-shot classification tasks with the CLIP model.
Specifically, Block-LoRA firstly partitions the original low-rank decomposition matrix of LoRA into a series of sub-matrices, and then shares all down-projection sub-matrices to reduce redundancy.
These operations offers two key advantages: (1) it reduces the number of parameters that need to be trained; (2) during the forward propagation, we transforms a part of complex matrix multiplication operations in vanilla LoRA into simpler matrix additions, thereby significantly reducing the computational cost of fine-tuning. 
Consequently, Block-LoRA enables fine-tuning CLIP on the ImageNet Few-Shot benchmark using a single 24GB GPU. 
Moreover,  Block-LoRA  inherits the key benefits of the vanilla LoRA, including: (1) ease of switching between different tasks. It can be achieved by simply replacing the low-rank submatrices according to tasks without modifying the original pre-trained model; (2) no additional inference latency. During inference, the introduced low-rank submatrices are all merged into the model’s original weights, thus without adding extra computational overhead and maintaining inference efficiency.
Without bells and whistles, the experimental results on few-shot learning, cross-dataset evaluation, and domain generalization tasks demonstrate that the proposed Block-LoRA achieves competitive classification performance, compared to the existing SOTA few-shot methods, while significantly reducing trainable parameters and computational overhead.
In summary, our key contributions of this work include: (1) To our knowledge, this is the first study to reveal redundancy in the vanilla LoRA structure for few-shot image classification tasks. (2) We propose a novel Block Matrix based Low Rank Adaptation framework, Block-LoRA, which utilize matrix partitioning and shared down-projection sub-matrices to reduce the computational cost of fine-tuning. (3) Extensive experiments of few-shot classification on 11 datasets demonstrate that our proposed Block Matrix achieve competitive performance compared to SOTA methods, while maintaining a low training parameters count and reduced computational costs.

    


\section{Related Work}
\subsection{Vision-Language Models}
Vision-Language Models (VLMs) have gained significant attention in recent years \cite{zhang2024vision},  driving substantial advancements in integrating visual and textual data. 
A typical example is CLIP (Contrastive Language-Image Pretraining) \cite{radford2021learning}, which employs a contrastive learning approach to jointly train an image encoder and a text encoder. CLIP learns to align the features of images and their corresponding text descriptions feature, enabling zero-shot transfer to various downstream tasks by comparing the similarity between between encoded image and text representations.
Based on CLIP, ALIGN \cite{jia2021scaling} leverage a large-scale noisy dataset without requiring expensive filtering or post-processing steps during the data-gathering phase.   
By training on massive multi-modal datasets in an end-to-end manner, VLMs have  demonstrated remarkable performances on various downstream tasks by effectively leveraging pre-trained knowledge \cite{gu2023systematic,zhang2024vision}.  
However, despite their strong transfer capabilities, efficiently adapting VLMs to few-shot learning tasks while keeping training costs manageable remains a significant challenge.


\subsection{Few-Shot Learning}
Few-shot learning \cite{fei2006one,lake2015human,wang2020generalizing,lake2023human} aims to enable models to learn novel concepts from only using the limited labeled samples, mimicking human-like learning abilities.
Conventional few-shot learning methods \cite{vinyals2016matching,snell2017prototypical,zhu2023transductive,zhouless,Meta-AdaM,zhang2024metadiff} typically rely on an additional base dataset and the meta-training strategy to pre-train the models for the generalization capabilities \cite{MAML}. 
Recent advancements in pre-trained VLMs have led to a new paradigm in few-shot learning, where meta-training and base datasets are no longer required. 
For example, CoOP\cite{zhou2022learning} directly fine-tunes CLIP with the learnable context prompts to improve few-shot performance.
CoCoOP \cite{zhou2022conditional} extends CoOp by incorporating the input-conditional tokens, combined with the learnable context vectors. 
MaPle \cite{khattak2023maple} and PromptSRC\cite{khattak2023self} both employ the multi-modal prompt learning, applying prompting techniques to both the image and text encoder to enhance few-shot adaptation.
Another line of research focuses on adapter-based techniques \cite{houlsby2019parameter} to fine-tune CLIP for few-shot tasks. 
For instance, CLIP-Adapter \cite{Clip-adapter} integrates a learnable adapter with a bottleneck layer to refine feature transformations.
GraphAdapter\cite{graphadapter} introduces a graph learning based adapter to capture the structure knowledge.
More recently, CLIP-LoRA \cite{zanella2024low} directly applies LoRA to fine-tune CLIP. 
Despite the promising results achieved by these methods, most suffer from high computational costs and large parameter counts, making efficient fine-tuning of VLMs for few-shot tasks an ongoing challenge.

\section{Preliminaries}

\subsection{Contrastive vision-language pretraining}

Contrastive vision-language pre-training \cite{radford2021learning}  has demonstrated impressive transfer performance surpassing single-modal pre-training. A widely adopted model in this paradigm is CLIP \cite{radford2021learning} model, which is trained on approximately 400 million image-text pairs. CLIP consists of both an image encoder and a text encoder simultaneously to capture visual and textual features, encouraging the corresponding image and text features of a sample has the similar embedding. 
Formally, we denote the image encoder and text encoder as $f(\  )$ and $g(\  )$, respectively. Given a batch of image-text pairs $\{(x_i,t_i)\}$ from the training set, the extracted visual features and textual features from CLIP are :$ \boldsymbol{v_i} = f(x_i)$ and  $ \boldsymbol{t_i} = g(t_i)$. The contrastive loss of CLIP is computed using a similarity-based softmax function:
\begin{equation}
\mathcal{L}_{\mathrm{CE}}=-\sum_{ i \in Batch } \log \frac{\exp \left(\operatorname{sim}\left(\boldsymbol{v_i}, \boldsymbol{t_i} \right) / 
\lambda\right)}{\sum_{j \in Batch} \exp \left(\operatorname{sim}\left(\boldsymbol{v_i}, \boldsymbol{t_j} \right) / \lambda\right)},
\end{equation}
where $Batch$ denotes the number of image-text pairs in the batch, $\operatorname{sim}$ represents  cosine similarity, and  $\lambda$  indicates the temperature parameter.
During inference, CLIP enables zero-shot classification task to classify an unlabeled image $x$ into one of K possible categories without training. And K pre-defined textual prompts $\{t_i\}_{i=1}^{K}$ (e.g., ``a photo of a [class]") are fed into the textual encoder $g(\  )$. The prediction of $x$ is formulated as:
\begin{equation}
p(y = i | x)  =  \frac{\exp \left(\operatorname{sim}\left(\boldsymbol{v_i}, \boldsymbol{t_i}  \right) / \lambda\right)}{\sum_{j \in K} \exp \left(\operatorname{sim}\left(\boldsymbol{v_i}, \boldsymbol{t_j} \right) / \lambda\right)}
\end{equation}

\subsection{CLIP-based Few-Shot Learning}
Although CLIP enable zero-shot Classification without training, its performance degrades significantly in the presence of domain shifts or uncertain
concepts in downstream tasks.
Consequently, fine-tuning CLIP with few available samples from downstream datasets is vital.  
Formally, in CLIP-based Few-Shot Learning, a collection of known images $\{x_i| i \in NK \}$ from a downstream dataset is given, where $N$ denotes the total number of categories and $K$ is the number of labeled images (shot) per category, which is typically small (e.g., $K \le 16$). The goal of few-shot learning is to predicted the labels of the remained unknown images in the dataset, using only these $NK$ samples, a setting commonly referred to as K-shot learning.
Inspired by recent progresses of Parameter-Efficient Fine-Tuning (PEFT) in NLP \cite{ding2023parameter,han2024parameter}, most proposed CLIP-based few-shot learning methods utilize the PEFT paradigms to fine-tune the pretrained CLIP, showing better performance and parameter-efficient compared to the full fine-tuning. They will only fine-tune a small portion of parameters in CLIP or introduce the additional learnable modules, while keeping the most portion of parameters in CLIP frozen. 
And the fine-tuned visual feature and textual feature with few-shot data are denoted as $\boldsymbol{\tilde{v}_i}$ and $\boldsymbol{\tilde{t}_i}$, respectively.
And the training process of PEFT with CLIP is formulated as $\boldsymbol{\tilde{v}_i}$ and $\boldsymbol{\tilde{t}_i}$, respectively. The training process  is formulated as:
\begin{equation}
\mathcal{L}_{\mathrm{FSL}}=
-\sum_{ i \in NK } \log \frac{\exp \left(\operatorname{sim}\left(\boldsymbol{\tilde{v}_i}, \boldsymbol{\tilde{t}_i} ) \right) / 
\lambda\right)}{\sum_{j=1}^{NK}  \exp \left(\operatorname{sim}\left(\boldsymbol{\tilde{v}_i}, \boldsymbol{\tilde{t}_j} )\right) / \lambda\right)},
\end{equation}

\section{Method}
In this section, we first introduce the Low-Rank Adaptation (LoRA), and then details the Block Matrix based Low-Rank Adaptation (Block-LoRA).

\begin{figure*}[t]
	\centering
	\includegraphics[width=1.9\columnwidth]{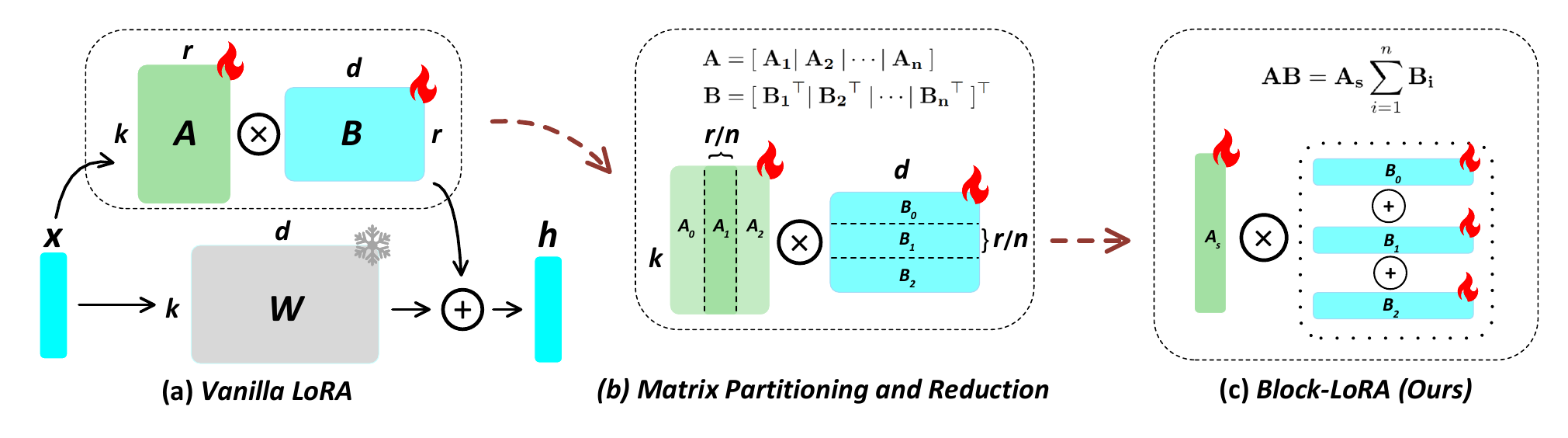}
	\caption{The detail structure of our proposed Block-LoRA.}
	\label{fig:block}
\end{figure*}

\subsection{Low-Rank Adaptation (LoRA) }
Low-Rank Adaptation (LoRA) \cite{hu2022lora} is a recently proposed PEFT method, which simulates the update of the pre-trained model weights by injecting trainable low-rank decomposition matrices into each layer, while  keep the original model weights frozen. 
As illustrated in Fig \ref{fig:block} (a), LoRA only optimizing the low-rank decomposition matrices, when adapting to a downstream task.
Specifically, given a pre-trained weight $\mathbf{W} \in \mathbb{R}^{k \times d} $,  the updated weight matrix $\mathbf{\hat{W}}$ by LoRA is formulated as: 
$\mathbf{\hat{W}} = \mathbf{W} +  \Delta \mathbf{W} =  \mathbf{W} + \mathbf{BA} $, where a low-rank decomposition $\Delta \mathbf{W} = \mathbf{AB}, \mathbf{A} \in \mathbb{R}^{k \times r}, \mathbf{B} \in \mathbb{R}^{r \times d }$ models this weight update during training, and the rank $r \ll min(d,k)$. Consequently, the forward pass with LoRA is formulated as: 
\begin{equation}\label{eq:4}
h  =  x\mathbf{W}  +  x \Delta \mathbf{W}  =  x \mathbf{W} + x \mathbf{AB} 
\end{equation}
where the original pre-trained weight $\mathbf{W}$ remains frozen, while two low-rank  decomposition matrices $\mathbf{A}$ and $\mathbf{B}$ are learnable during training. The recent CLIP-LoRA \cite{zanella2024low} directly applies LoRA to the image and text encoder of CLIP for improving few-shot learning.

\subsection{Block Matrix based Low-Rank Adaptation}
Though using LoRA to fine-tune CLIP achieves the promising few-shot performance, it ignore the existed redundancy in the vanilla LoRA structure. 
To alleviate this, we introduce Block-LoRA to further reduce the redundancy for few-shot adaptation while maintaining effectiveness.
Similar to LoRA, we assume the low-rank parameter update follows: $\mathbf{\hat{W}} = \mathbf{W} +  \Delta \mathbf{W} =  \mathbf{W} + \mathbf{AB} $. 
As detailed in Fig \ref{fig:block} (b), the proposed Block-LoRA first partitions the two vanilla low-rank matrices $\mathbf{A} \in \mathbb{R}^{k \times r}$ and  $\mathbf{B} \in \mathbb{R}^{r \times d }$ into multiple submatrix multiplications along the dimension of rank $r$, without modifying other dimensions:
\begin{align}
\mathbf{A} = \ &[\  \mathbf{A_1} |\  \mathbf{A_2}\  | \cdots |\   \mathbf{A_n}\   ] \\
\mathbf{B} = \ &[\  \mathbf{B_1} ^\top |\  \mathbf{B_2} ^\top\  | \cdots |\   \mathbf{B_n}^\top \   ]^\top
\end{align}
where each submatrix $\mathbf{A_i} \in \mathbb{R}^{k  \times \frac{r}{n}}$ and  $\mathbf{B_i} \in \mathbb{R}^{\frac{r}{n}  \times d}$ represents the $i$-th block matrices (or submatrices) of $\mathbf{A}$ and $\mathbf{B}$, respectively. And $n$ denotes the number of blocks in a matrix. $\mathbf{B_i}^\top$ means denotes the matrix transposition operation. $|$ denotes that submatrices are concatenated column-wise or second dimension.
Since the second dimension of $\mathbf{A}$ and the first dimension of $\mathbf{B}$ have the same dimension $\frac{r}{n}$, and we deduce that:
\begin{equation}
\begin{aligned}
\mathbf{AB} =  &[\  \mathbf{A_1} |\  \mathbf{A_2}\  | \cdots |\   \mathbf{A_n}\   ]  *  [\  \mathbf{B_1} ^\top |\  \mathbf{B_2} ^\top\  | \cdots |\   \mathbf{B_n}^\top \   ]^\top \\
= & \sum_{i=1}^{n}  (\mathbf{A_i}  \mathbf{B_i} ) \in  \mathbb{R}^{k \times d} 
\end{aligned}
\end{equation}
Substituting this into Eq (\ref{eq:4}), the forward pass becomes:
\begin{equation}\label{eq:8}
\begin{aligned}
h  = & \ x \mathbf{W} + x \mathbf{AB} =  x \mathbf{W} + x \sum_{i=1}^{n}  (\mathbf{A_i}  \mathbf{B_i} )
\end{aligned}
\end{equation}

Prior studies indicate that the dimensionality reduction matrix $\mathbf{A}$ of vanilla LoRA structure typical contains redundancy \cite{zhu2024asymmetry,zhang2023lora}. However, simply freezing $\mathbf{A}$ after random initialization degrade performance in certain downstream tasks. 
As illustrated in Fig \ref{fig:block} (c), to balance efficiency and performance, we deploy a shared down-projection matrix $\mathbf{A_s} \in  \mathbb{R}^{k  \times \frac{r}{n} }$ to replace all block matrices $\mathbf{A_i}$ in Eq (\ref{eq:8}). 
This operation not only avoids the performance degradation, but also reduces the redundancy of matrix $\mathbf{A}$ in the vanilla LoRA.
Formally, we have:
\begin{equation}\label{eq:9}
h  =  \  x \mathbf{W} + x \sum_{i=1}^{n}  (\mathbf{A_s}  \mathbf{B_i} ) 
=  \  x \mathbf{W} + x \mathbf{A_s} \sum_{i=1}^{n}    \mathbf{B_i}  
\end{equation}
We denote this approach as Block-LoRA$(r,n)$, where $r$ denotes the rank of the low-rank  decomposition, and $n$ is the number of blocks. 

In summary, our proposed Block-LoRA has following advantages: (1) Block-LoRA lowers the number of training parameters compared to vanilla LoRA, without compromising few-shot performance. Block-LoRA use the a shared down-projection matrix $\mathbf{A_s}  \in  \mathbb{R}^{k  \times \frac{r}{n} } $ to replace the vanilla $\mathbf{A}  \in \mathbb{R}^{k  \times r} $, the parameter count of which is reduced by a factor of $\frac{1}{n}$, effectively mitigating redundancy of $\mathbf{A}$ and overfitting.
(2) Block-LoRA reduces the computational complexity compared to vanilla LoRA. As described in Eq (\ref{eq:9}), the dimension of the matrix used in Block-LoRA is lower than before, such as  $\mathbf{A_s}$ and $\mathbf{A}$. Thus, Block-LoRA transforms a part complex matrix multiplication operations $\mathbf{AB}$ in vanilla LoRA into more efficient matrix operations $\mathbf{A_s} \sum_{i=1}^{n}    \mathbf{B_i}$, thereby significantly lowering the computational cost of fine-tuning. 
(3) Block-LoRA inherits the advantages of vanilla LoRA, which won't introduce the additional inference latency, unlike the previous adapter-based or prompt-based methods. The update matrix $\Delta \mathbf{W}$ is merged into the original weights $\mathbf{W}$ during inference, ensuring no extra computational head, which does not affect the inference speed.
Moreover, Block-LoRA facilitates seamless task switching by simply replacing low-rank submatrices, without modifying the pre-trained model.


\begin{figure*}[t]
	\centering
	\subfloat{
		\includegraphics[scale=0.3]{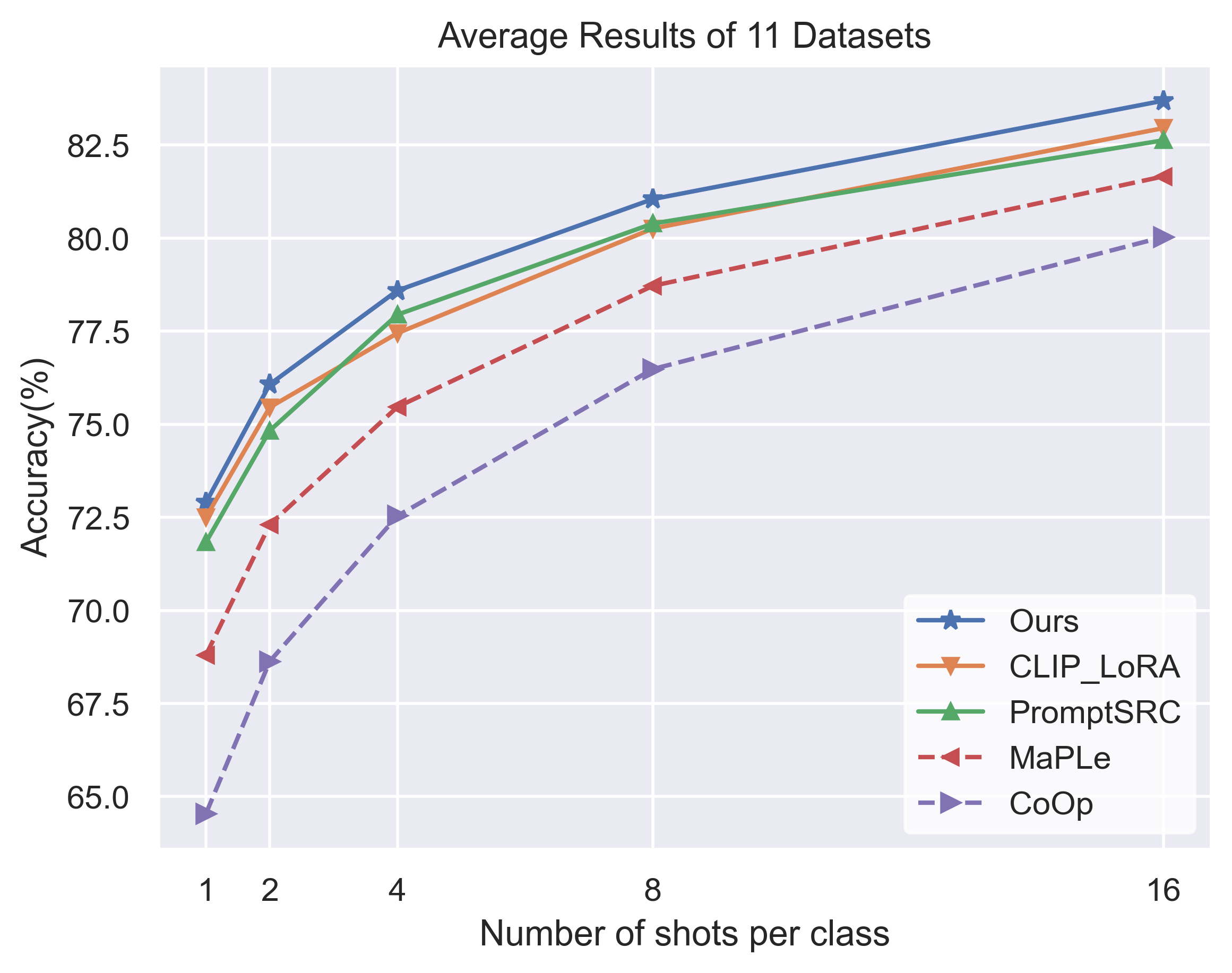}
		\includegraphics[scale=0.3]{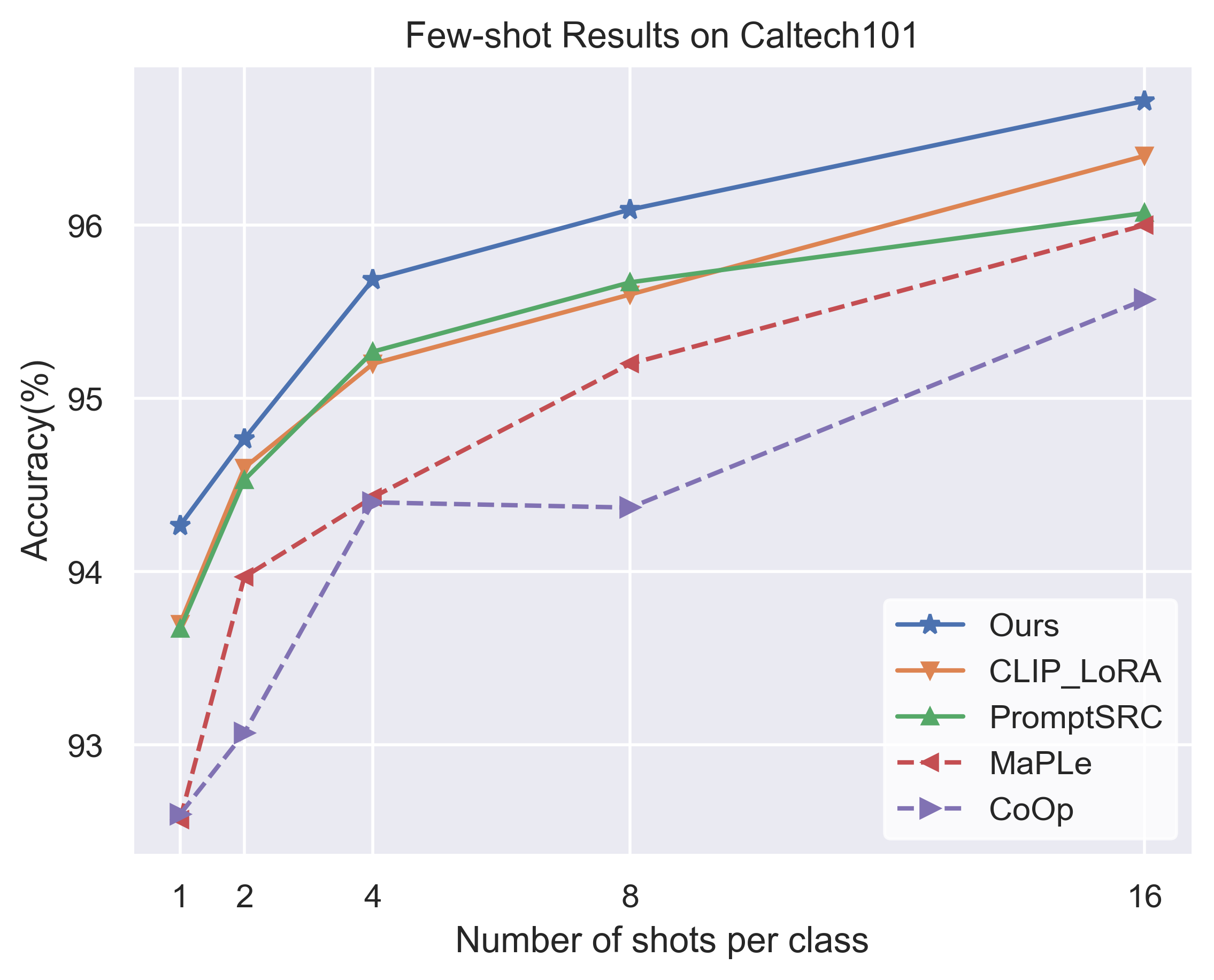}
		\includegraphics[scale=0.3]{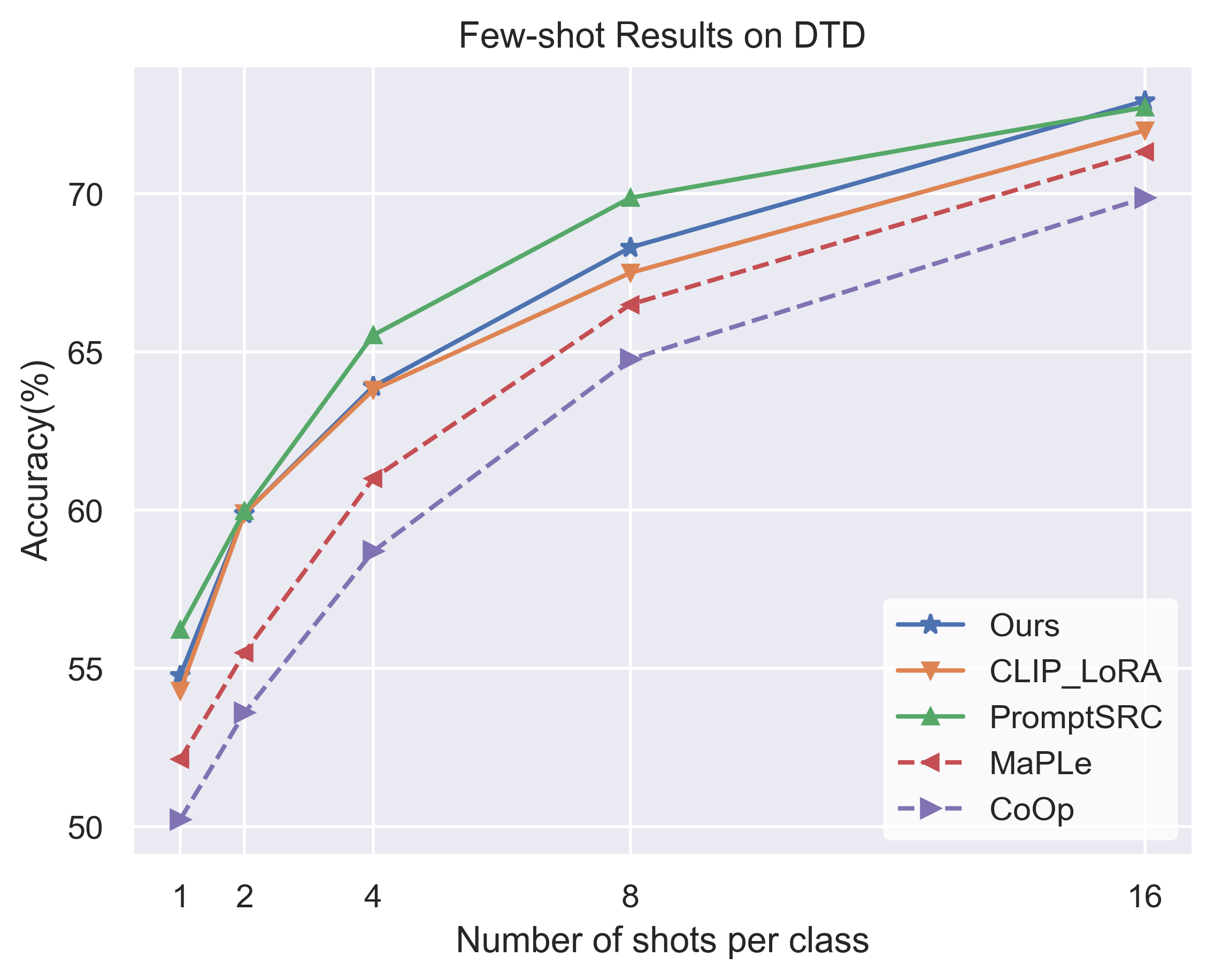}
            \includegraphics[scale=0.3]{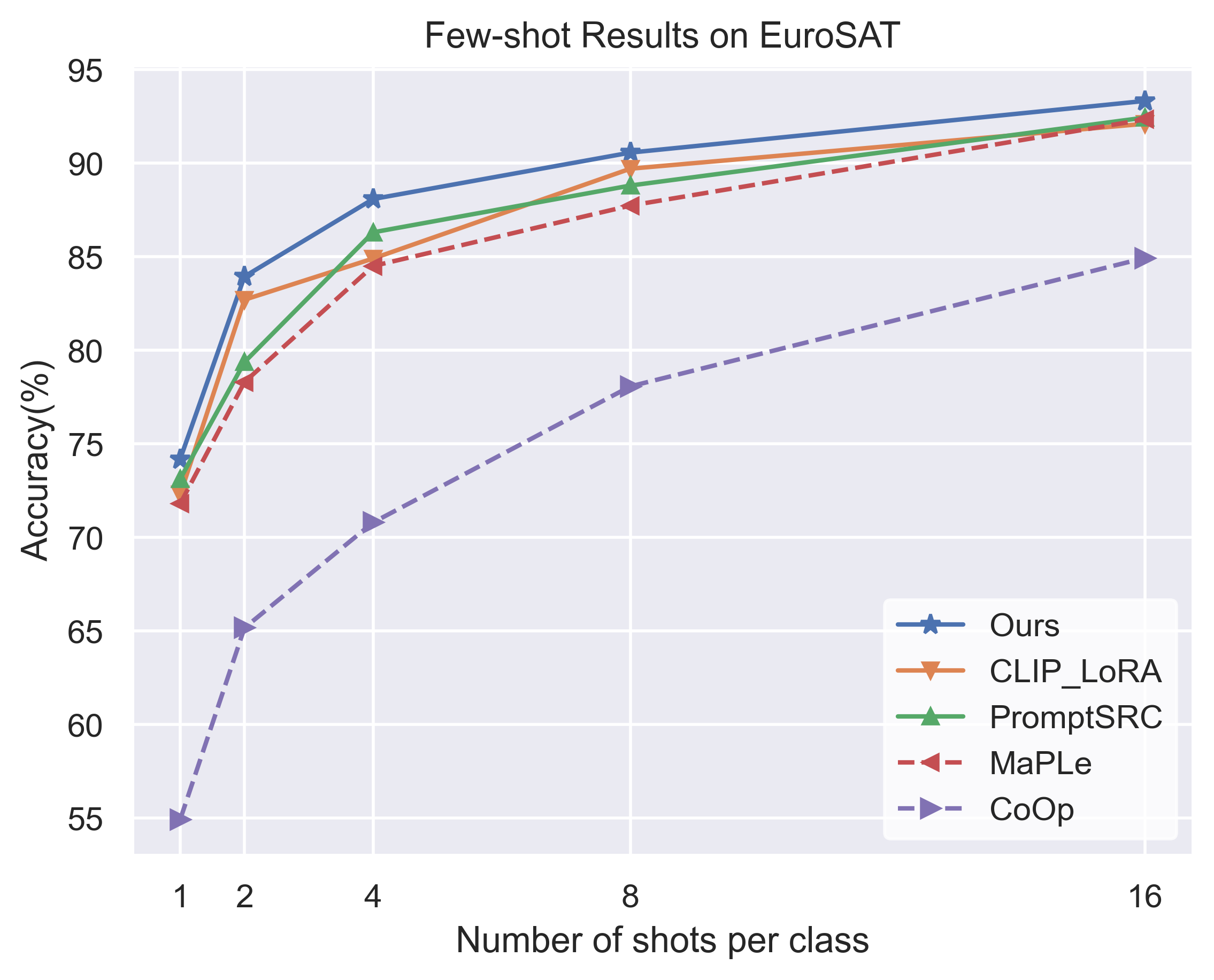}
        }\\
	\subfloat{
		\includegraphics[scale=0.3]{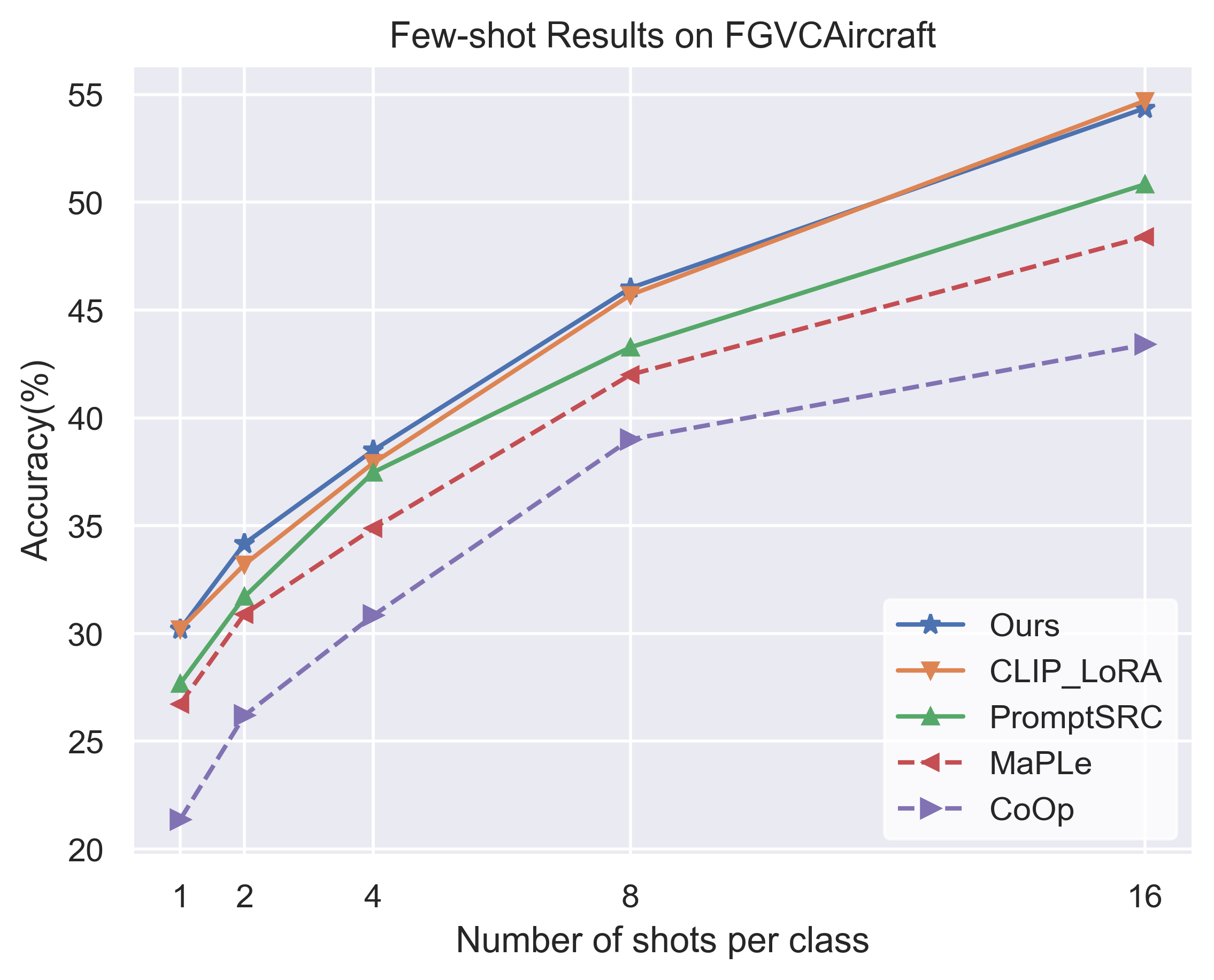}
		\includegraphics[scale=0.3]{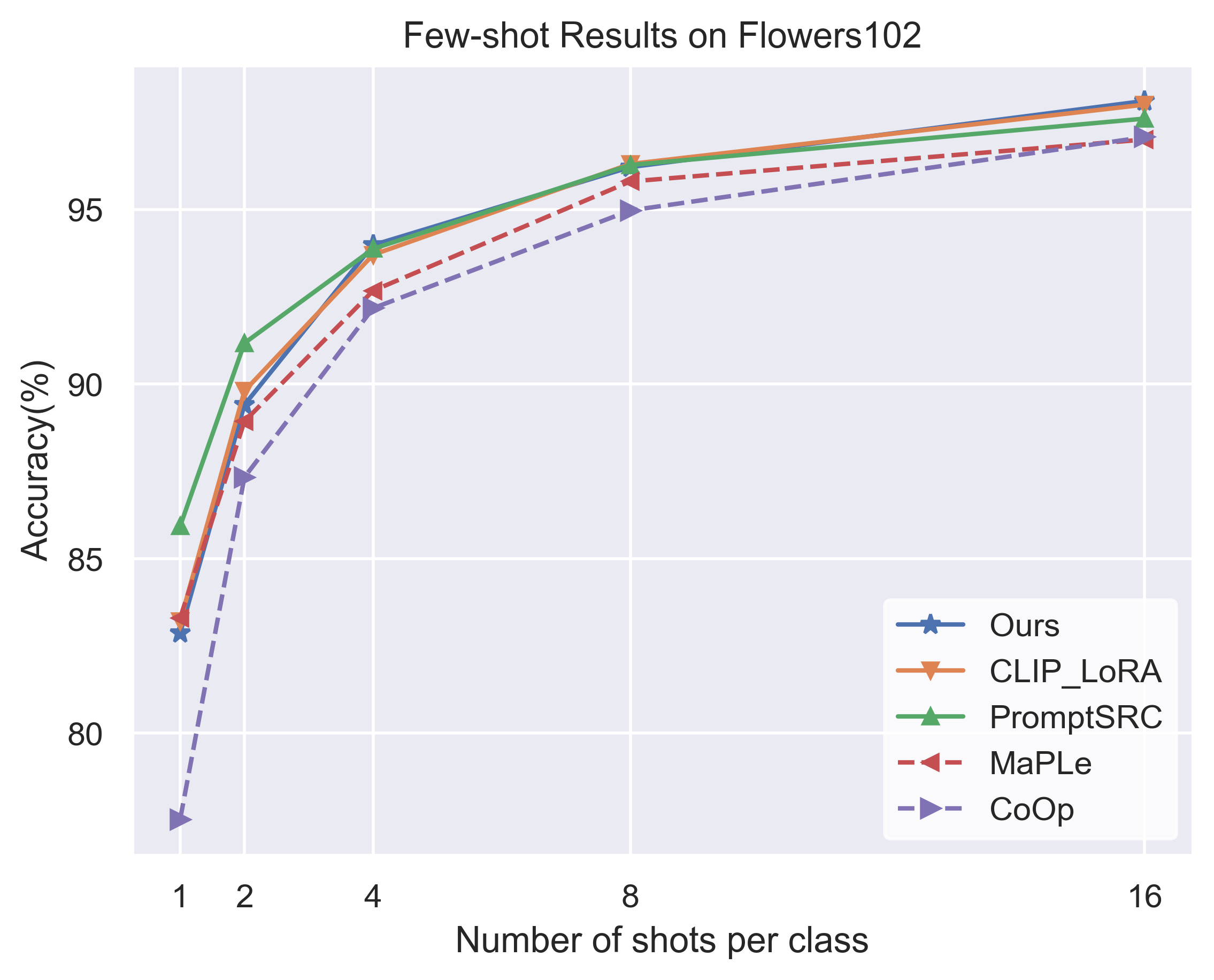} 
            \includegraphics[scale=0.3]{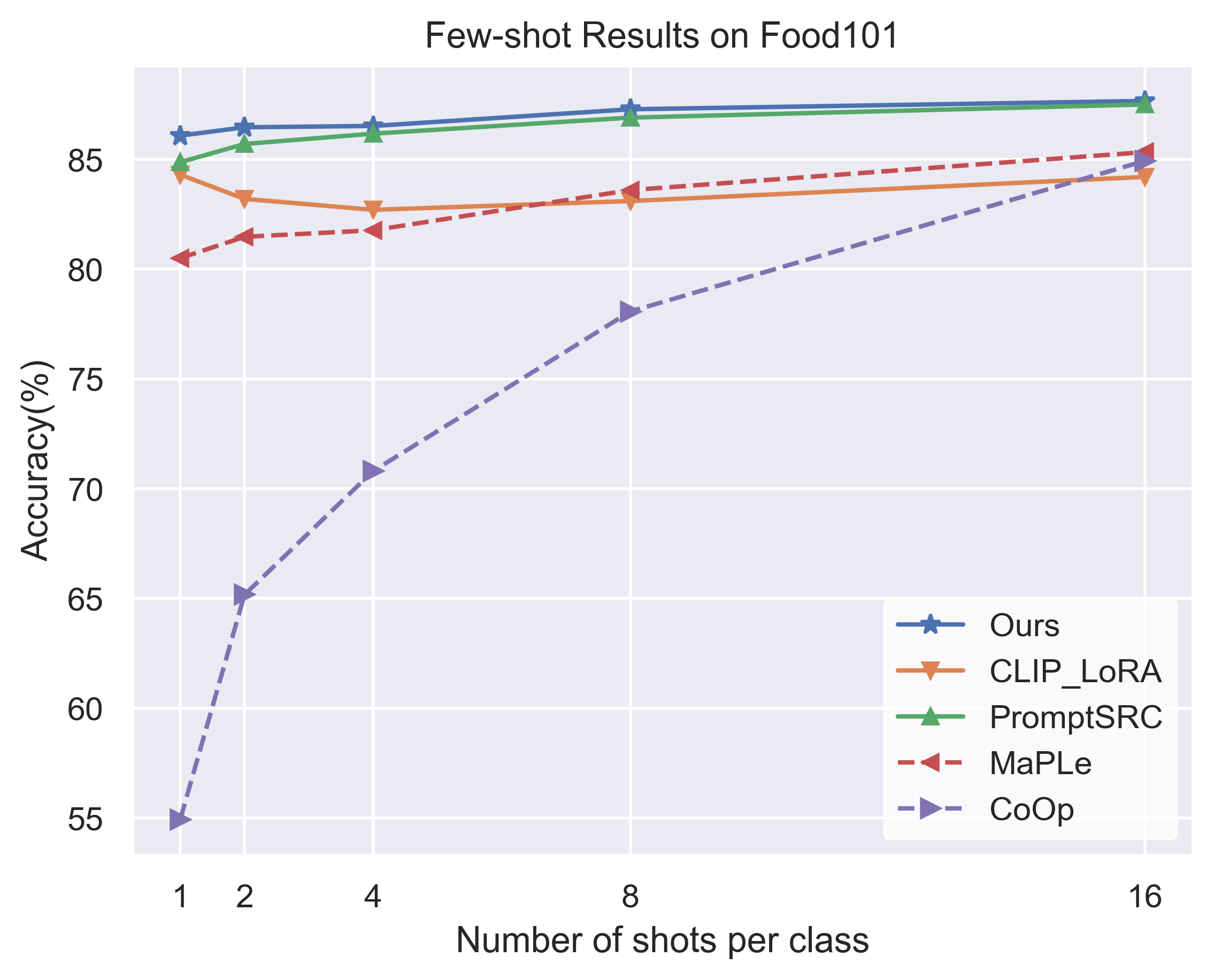}
		\includegraphics[scale=0.3]{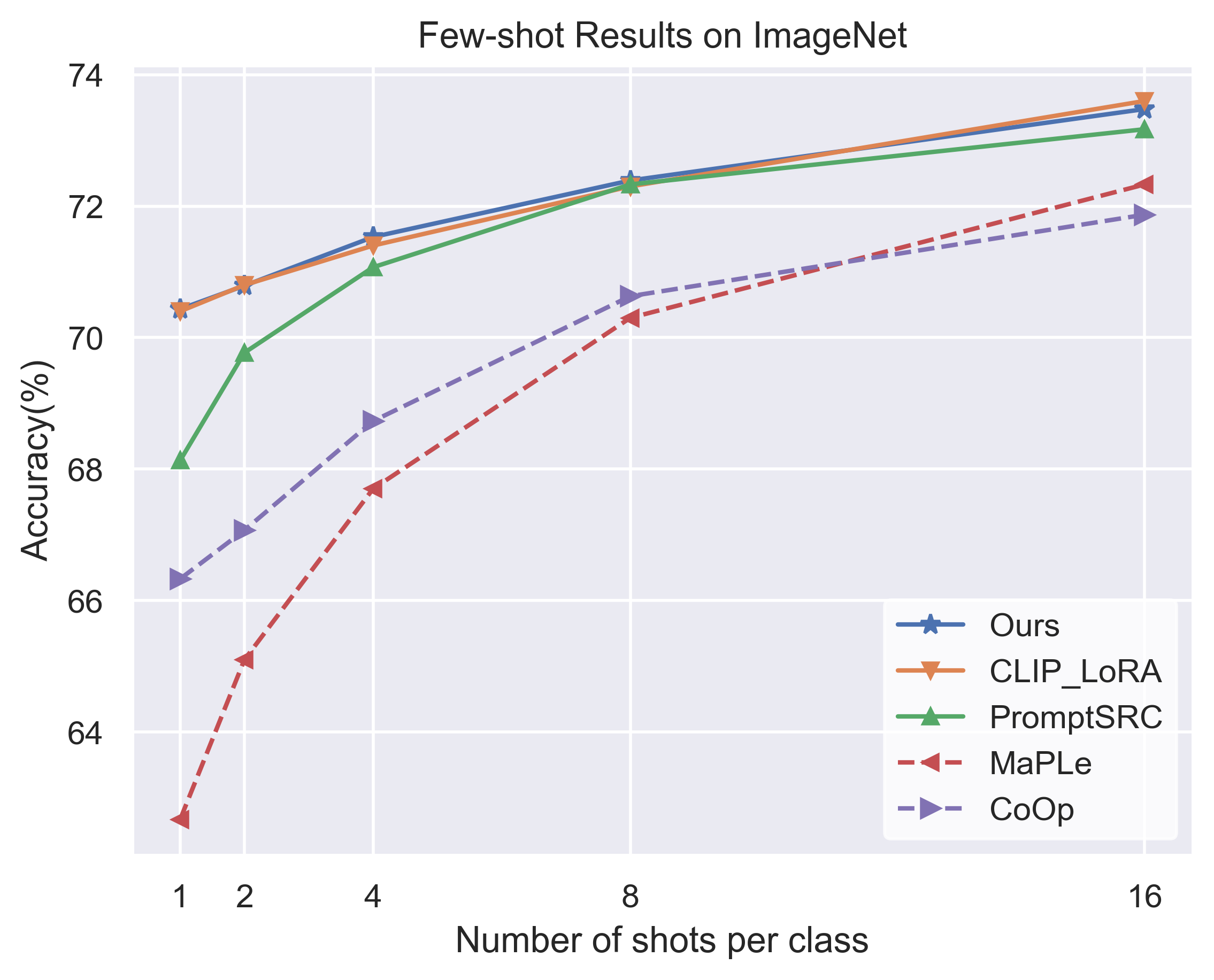}
        } \\
	\subfloat{
		\includegraphics[scale=0.3]{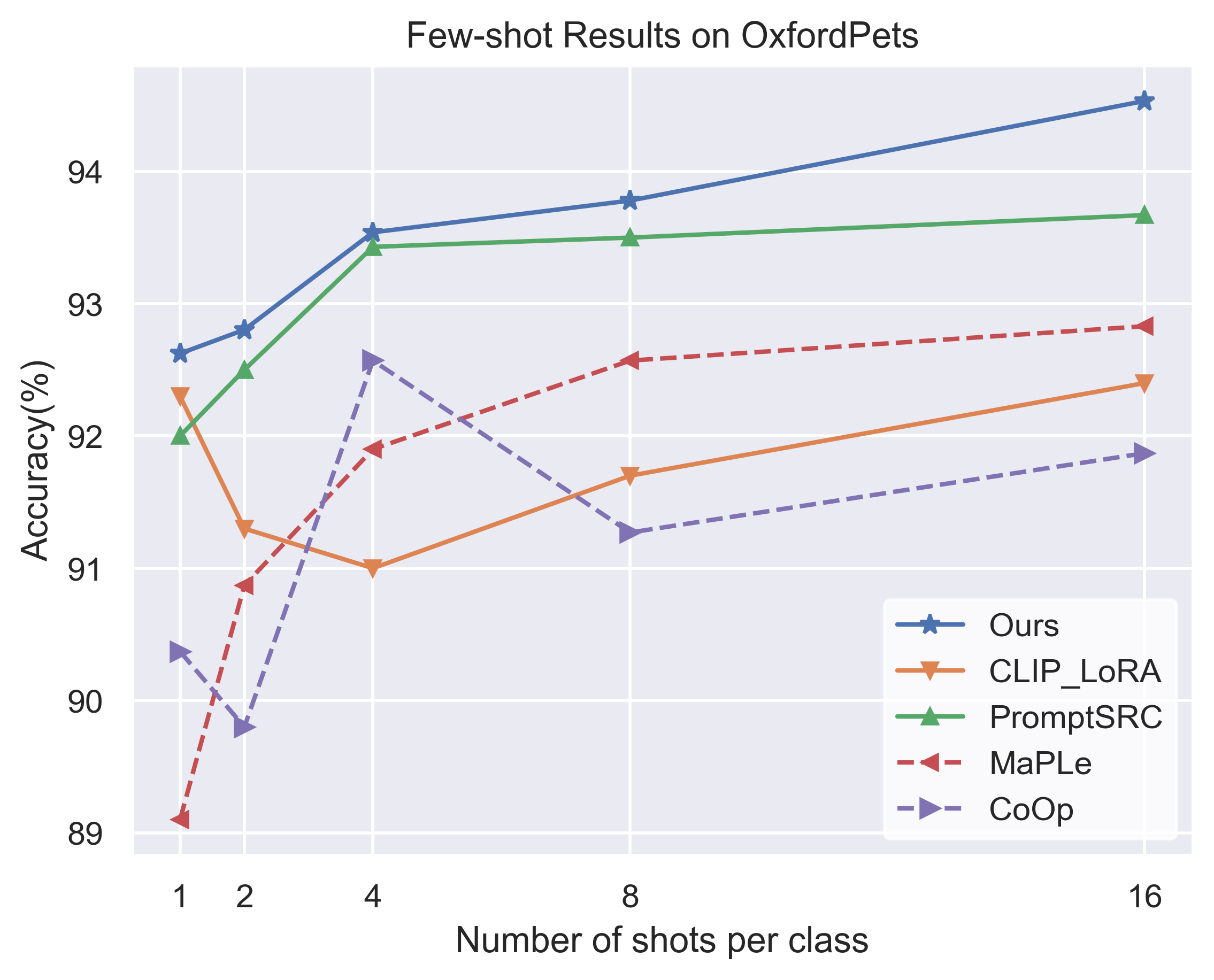} 
		\includegraphics[scale=0.3]{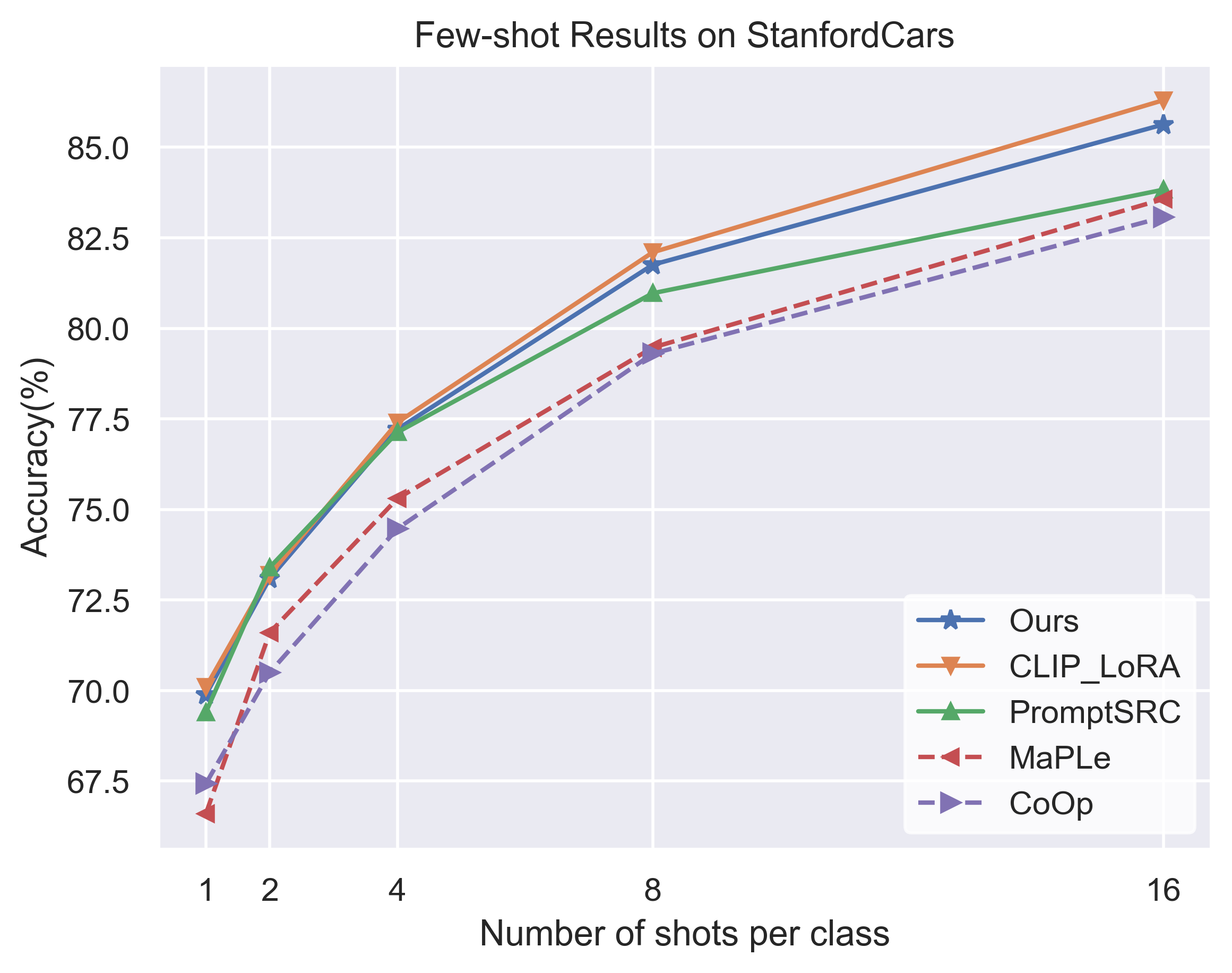}
		\includegraphics[scale=0.3]{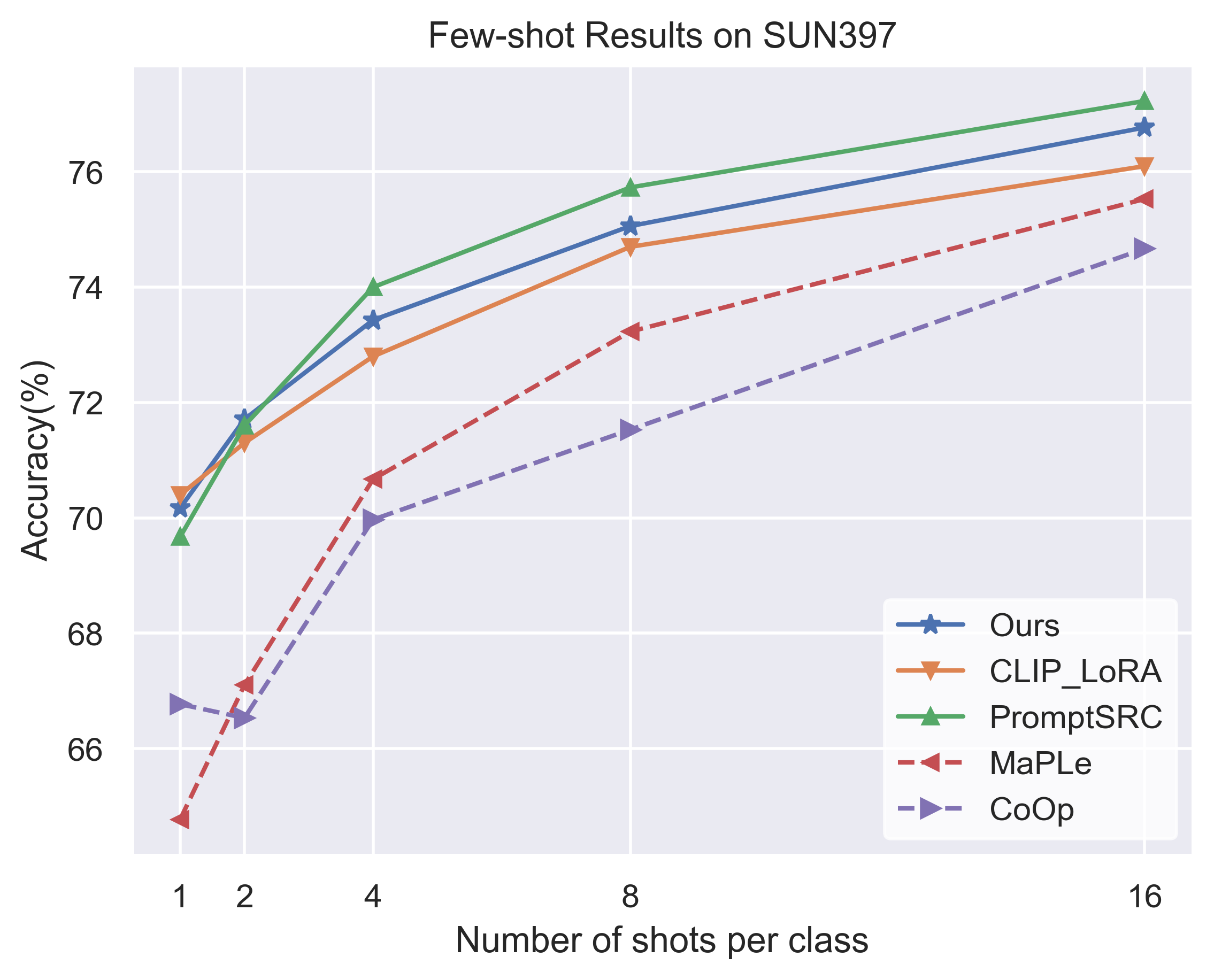} 
		\includegraphics[scale=0.3]{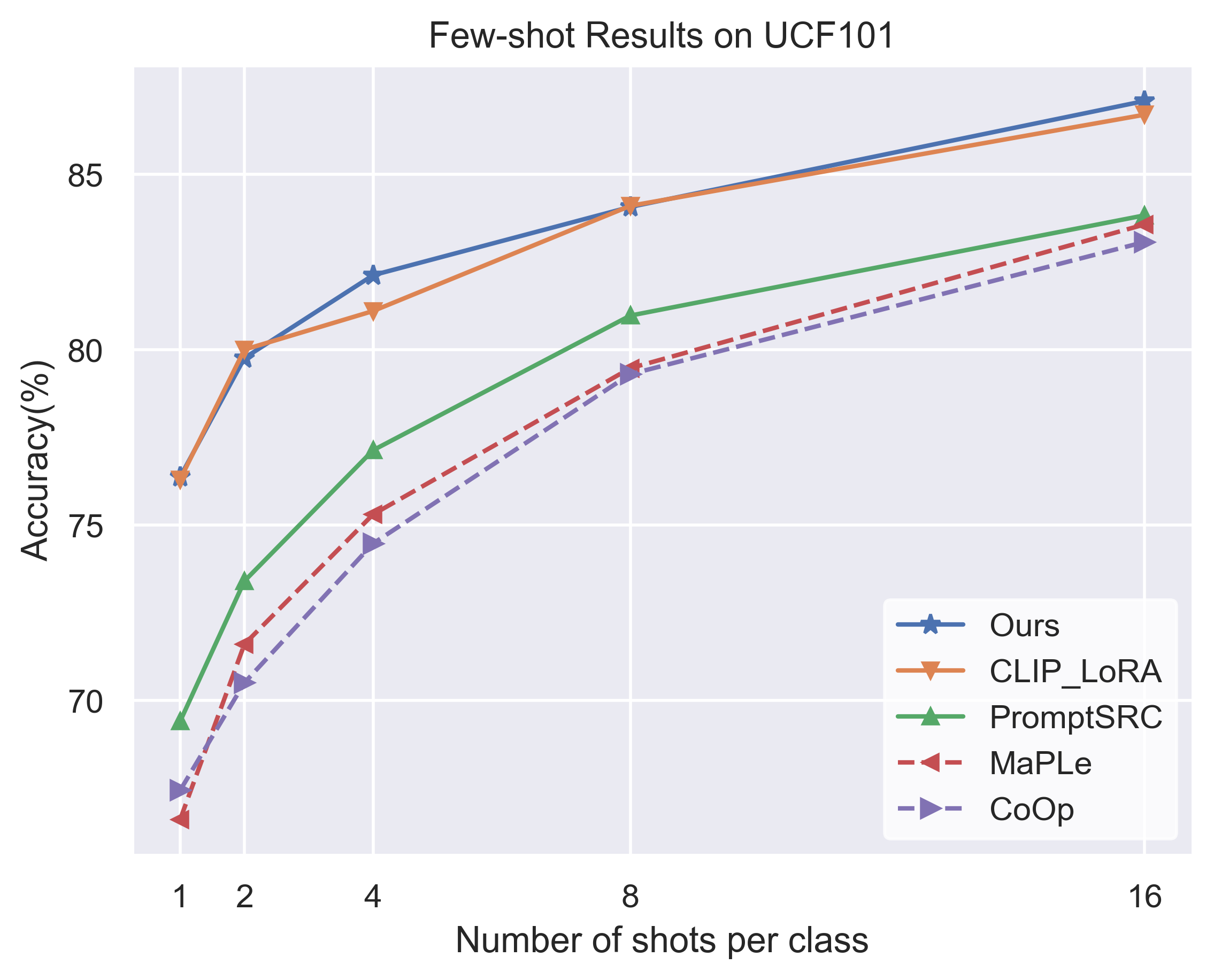} }
	\caption{Block-LoRA performance comparison in few-shot classification tasks.}
	\label{fig:few}
    \vspace{-0.4cm}
\end{figure*}

\subsection{Theoretical Analysis}

We define $\boldsymbol{W}=\{\mathbf{W}_{l}\}_{l = 1}^{L}$ as the all \(L\) parameter matrices of a pre-trained model. Let \(I\subseteq\{1,\cdots,L\}\) be a set of the parameter matrice index to be fine-tuned. Given a labeled training set \(S \in Z \) with  i.i.d. training examples from unknown real data distribution $\mu$, we analyze the adaptation process. 
Let \(r\) denote the rank in both LoRA and Block-LoRA. We define the number of training as $\#S$ and suppose each tuned parameter is quantized to \(q\) bits. We define the fine-tuning based adaptation frameworks by using an adaptation matrices set as \(\Delta \boldsymbol{W} =\{\Delta \mathbf{W}_{l}\}_{l = 1}^{L}\), 
leading to the fine-tuned parameters: 
$\boldsymbol{\hat{W}} =\{ \mathbf{\hat{W}_=l} \}_{l = 1}^{L}$ 
with \(\mathbf{\hat{W}}_{l}=\mathbf{W}_{l} +\Delta \mathbf{W}_{l} \) for \(l\in I\). 
We formally define the LoRA and Block-LoRA adaptation as follows:
\begin{itemize}
\item LoRA: For each \(l\in I\), enforce \(\Delta \mathbf{W}_{l}=\mathbf{A}_{l} \mathbf{B}_{l}\) and optimize \(\{\mathbf{A}_{l},\mathbf{B}_{l}\}_{l\in I}\) to fit the training data \(S\), as described in Eq (\ref{eq:4}).

\item Block-LoRA: For each \(l\in I\), enforce \(\Delta \mathbf{W}_{l}=\mathbf{A_s}_l \sum_{i=1}^{n}   \mathbf{B_i}_l \) and optimize \(\{ \mathbf{A_s}_l ,\sum_{i=1}^{n}   \mathbf{B_i}_l \}_{l\in I}\) to fit the training data \(S\), as described in Eq (\ref{eq:8}).
\end{itemize}
We denote the generalization error of an algorithm as $error\left(\mu,\text{algorithm}\right) $ \cite{zhu2024asymmetry}. The following lemma provides an upper bound on this generalization error:

\noindent \textbf{Lemma 1.}\ 
Assume that the loss \(\ell^{\boldsymbol{W}}(\Delta \boldsymbol{W},Z)\) is \(\sigma\)-sub-Gaussian under \((\Delta \boldsymbol{W},Z)
\sim P_{\Delta \boldsymbol{W} |\boldsymbol{W} }\times\mu\). Then,
\begin{align*}
\left|error\left(\mu,\text{LoRA}\right)\right|&\leq\sqrt{\frac{2rq\sigma^{2}\ln 2}{\#S}\sum_{l\in\mathcal{I}}\left(k_{(l)} + d_{(l)} \right)}\\
\left|error\left(\mu,\text{Block-LoRA}\right)\right|&\leq\sqrt{\frac{2rq\sigma^{2}\ln 2}{\#S}\sum_{l\in\mathcal{I}}\left(\frac{k_{(l)}}{n} + d_{(l)} \right)}
\end{align*}
where $k_{(l)}$ and $d_{(l)}$ denote the input and output feature dimensions of the $l$-th parameter matrix. 
From the above bounds, we observe that Block-LoRA provides a tighter generalization upper bound compared to LoRA, despite having fewer trainable parameters. The full proof is provided in the Appendix.



\begin{table*}[t] \scriptsize
	\centering
	\caption{Cross-dataset benchmark evaluation.}
	\setlength{\tabcolsep}{3.3mm}{
	\begin{tabular}{l | c | cccccccccc c}
		\hline
		& Source & \multicolumn{10}{c}{Target} &  \\
		Methods &\rotatebox{72}{ImageNet } & \rotatebox{72}{Caltech101} & \rotatebox{72}{OxfordPets} & \rotatebox{72}{StanfordCars }& \rotatebox{72}{Flowers102} & \rotatebox{72}{Food101} & \rotatebox{72}{Aircraft} & \rotatebox{72}{SUN397} & \rotatebox{72}{DTD}& \rotatebox{72}{EuroSAT} & \rotatebox{72}{UCF101} & \rotatebox{72}{Average} \\
		\hline
		CoOp & 71.51 & 93.70 & 89.14 & 64.51 & 68.71 & 85.30 & 18.47 & 64.15 & 41.92 & 46.39 & 66.55 & 63.88 \\
		CoCoOp   & 71.02 & \textbf{94.43} & 90.14 & 65.32 & 71.88 & 86.06 & 22.94 & 67.36 & 45.73 & 45.37 & 68.21 & 65.74 \\
		MaPLe   & 70.72 & 93.53 & 90.49 & 65.57 & 72.23 & 86.20 & 24.74 & 67.01 & 46.49 & \textbf{48.06} & 68.69 & 66.30 \\
		PromptSRC  & 71.27 & 93.60 & 90.25 & 65.70 & 70.25 & 86.15 & 23.90 & 67.10 & 46.87 & 45.50 & 68.75 & 65.81 \\
		CLIP-LoRA  & 73.48 & 94.03 & \textbf{91.01} & 66.70 & 72.62 & \textbf{86.54} & \textbf{25.30} & 68.19 & \textbf{46.91} & 42.07 & \textbf{70.10} & 66.99 \\
		Ours  & \textbf{73.60} & 94.09 & 90.90 & \textbf{66.95} & \textbf{72.73} & 86.46 & 25.27 & \textbf{68.27} & 46.63 & 42.21 & 69.88 & 67.00 \\
		\hline
		Rank (Ours) & 1 & 2 & 2 & 1 & 1 & 2 & 2 & 1 & 3 & 5 & 2 & 1 \\
		\hline
	\end{tabular}
}
	\label{tab:cross}
\end{table*}

\section{Experiments}
\subsection{Datasets and Settings}
We evaluate our method on 11 publicly available datasets widely used in few-shot classification and cross-dataset transfer tasks, strictly following prior works \cite{zhou2022learning,khattak2023self,zanella2024low}. These datasets include: ImageNet, Caltech101, OxfordPets, StanfordCars, Flowers102, Food101, FGVCAircraft, SUN397, UCF101, DTD and EuroSAT. 
For domian generalization, we use 4 publicly available ImageNet variants, including: 
ImageNetV2, ImageNet-Sketch, ImageNet-A(dversarial) and ImageNet-R(endition). 
Detailed dataset statistics are provided in the appendix.

For a fair comparison, we adopt a ViT-B/16 \cite{dosovitskiy2020image} based CLIP
model in our experiments, consistent with prior studies \cite{zhou2022learning,khattak2023self,zanella2024low}. We use the official CLIP code and pre-trained weights, which are publicly available\footnote{https://github.com/openai/CLIP}.
All experiments were implemented using pytroch and repeated 3 times to report the average classification accuracy. 
Unless stated otherwise, we set the default rank $r =2 $ and the number of blocks $n=2$ in our proposed method, named as Block-LoRA$(2,2)$.
Additional results with varying hyperparameters are provided in the model analysis section.
We use the AdamW optimizer to train our model with 
a learning rate of $2e-4$ and a cosine annealing scheduler to adjust the learning rate dynamically during training.
Specially, the input text prompts we used is the simple ``a photo of a [class]" across all datasets, without complex manual prompt engineering.
Due to Block-LoRA's computational efficiency, even the experiments on the large ImageNet few-shot benchmark can run on a single GPU with 24GB memory.
All experiments in this paper are completed on an NVIDIA RTX 3090 GPU.

For few-shot classification tasks, all experiments are sampled from the training set of downstream dataset, and $K$ samples per class are randomly selected to train the model, where $K \in \{1, 2, 4, 8, 16\}$\cite{radford2021learning}. Subsequently, the trained model is used to predict the classification accuracy on the test set of the respective dataset.
For  cross-dataset transfer, the source domain is unified as 16-shot ImageNet \cite{zhou2022learning,khattak2023self}, and the target domian is the remaining 10 datasets. Notably, the model is only tested on the target domain datasets without further fine-tuning.
For domain generalization, the 16-shot ImageNet dataset serves as the source domain, while the four ImageNet variants (ImageNetV2, ImageNet-Sketch, ImageNet-A, and ImageNet-R) serve as target domains \cite{zhou2022learning,khattak2023self}. As in the transfer setting, the model is evaluated directly on the target domains without fine-tuning.

\subsection{Main Results}
\noindent\textbf{Few-shot Classification.}
This section mainly shows the results of the proposed Block-LoRA on few-shot image classification tasks, comparing to the recent methods, including CoOP \cite{zhou2022learning}, MaPLe \cite{khattak2023maple}, PromptSRC \cite{khattak2023self} and CLIP-LoRA \cite{zanella2024low}.
The few-shot image classification results on 11 datasets are summarized in Fig \ref{fig:few}.
Firstly, Block-LoRA demonstrates highly competitive few-shot performance across all 11 datasets. For example, Block-LoRA outperforms CLIP-LoRA in average accuracy on 11 datasets, across all few-shot settings (1, 2, 4, 8, and 16 shots). Notably, Block-LoRA has fewer parameters and lower computational overhead compared to CLIP-LoRA. Additionally, Block-LoRA also outperforms the previous SOTA PromptSRC \cite{khattak2023self} and MaPLe \cite{khattak2023maple}. For example, Block-LoRA achieves an average accuracy of $83.7\%$ on 11 datasets  in the 16-shot setting, which is a 1.3\% improvement over PromptSRC, despite PromptSRC employ text-based data augmentation enhance text information extraction \cite{khattak2023self}.
Secondly, for the individual dataset, Block-LoRA achieves a significant lead over other methods on EuroSAT and OxfordPets datasets.
For example, under the 4-shot setting on the EuroSAT dataset, Block-LoRA achieves $88.08\%$ accuracy, surpassing CLIP-LoRA by 3.7\%. 
Furthermore, for other datasets, Block-LoRA still achieves matching results, compared with other latest methods. For example, on the ImageNet dataset with 1000 categories, Block-LoRA achieves similar performance with LoRA-CLIP, while maintaining lower parameter count and computational complexity. Notably, Block-LoRA outperforms PromptSRC on the 1-shot ImageNet benchmark, achieving  $70.43\%$ accuracy, a $3.4\%$ improvement of  over PromptSRC.
Beyond accuracy,  Block-LoRA exhibits stable training behavior and efficient convergence, benefiting from its reduced parameter count. As the number of training samples increases (e.g., from 1-shot to 16-shot), Block-LoRA consistently improves downstream task performance. In contrast, the performance of  CoOP\cite{zhou2022learning} may be prone to oscillation, and in some cases, increasing the number of training samples can lead to degraded performance.

\noindent\textbf{Cross Dataset Evaluation.}
To further demonstrate the few-shot generalization capability of Block-LoRA, we conduct experiments on cross-dataset transfer tasks and compare Block-LoRA against several SOTA methods, including CoOP \cite{zhou2022learning}, CoCoOP\cite{zhou2022conditional}, MaPLe \cite{khattak2023maple}, PromptSRC \cite{khattak2023self} and CLIP-LoRA \cite{zanella2024low}. The results are summarized in Table \ref{tab:cross}. 
From these results, Block-LoRA demonstrates strong generalization across multiple datasets,  achieving the best performance on 4 datasets and the second-best performance on 5 datasets. 
Additionally, Block-LoRA’s performance is comparable to CLIP-LoRA in general, while maintaining lower computational complexity and fewer parameters.


\begin{table}[t] \scriptsize
	\centering
	\caption{Domain generalization evaluation.}
		\begin{tabular}{l |  c | ccccc}
			\hline
			& Source & \multicolumn{4}{c}{Target}    &       \\
			Methods & ImageNet       & \multicolumn{1}{c}{-V2} & \multicolumn{1}{c}{-S} & \multicolumn{1}{c}{-A} & \multicolumn{1}{c}{-R} & Avg.        \\
			\hline
			CLIP      & 66.73  & 60.83 & 46.15 & 47.77 & 73.96 & 57.18 \\
			CoOp      & 71.51  & 64.20 & 47.99 & 49.71 & 75.21 & 59.28 \\
			CoCoOp    & 71.02  & 64. 07 & 48.75 & 50.63 & 76.18 & 59.91 \\
			MaPLe     & 70.72  & 64.07 & 49.15 & 50.90 & 76.98 & 60.28 \\
			PromptSRC & 71.27  & 64.35 & 49.55 & 50.90 & 77.80 & 60.65 \\
			PBPrompt  & 71.71  & 64.53 & 49.32 & 51.64 & 76.71 & 60.55 \\
			CLIP-LoRA & 73.48          & \textbf{66.00}         & \textbf{50.07}        & 50.82                 & 77.50                  & \textbf{61.10 }         \\
			Ours & \textbf{73.60} & 65.89                  & 49.88                 & \textbf{50.88}        & \textbf{77.56}         & 61.05\\
			\hline
		\end{tabular}
	\label{tab:domain}
    \vspace{-0.4cm}
\end{table}

\noindent\textbf{Domain Generalization.}
To further assess Block-LoRA’s robustness under domain shifts, we evaluate its performance on domain generalization tasks, comparing to zero-shot CLIP (CLIP), CoOP \cite{zhou2022learning}, CoCoOP\cite{zhou2022conditional}, MaPLe \cite{khattak2023maple}, PromptSRC \cite{khattak2023self}, The results of PBPrompt \cite{liupatch} and CLIP-LoRA \cite{zanella2024low}. The results are reported in Table \ref{tab:domain}. From the results, Block-LoRA continues to demonstrate excellent  generalization performance, showing a certain degree of robustness, and achieving leading results on ImageNet-A and ImageNet-R. It also achieves second-best results on ImageNetV2 and ImageNet-S, which remains competitive with CLIP-LoRA while maintaining a more efficient parameterization.


\subsection{Model Analysis}

\begin{table*}[t] \scriptsize
	\centering
	\caption{ Performance effects of different rank $r$ and number of blocks $n$ in Block-LoRA.}
	\label{tab:ch5_rn}
	\setlength{\tabcolsep}{3.2mm}{
		\begin{tabular}{c|ccccccccccc|c}
			\hline
			$(r,n)$ &\rotatebox{72}{ImageNet } & \rotatebox{72}{Caltech101} & \rotatebox{72}{OxfordPets} & \rotatebox{72}{StanfordCars }& \rotatebox{72}{Flowers102} & \rotatebox{72}{Food101} & \rotatebox{72}{Aircraft} & \rotatebox{72}{SUN397} & \rotatebox{72}{DTD}& \rotatebox{72}{EuroSAT} & \rotatebox{72}{UCF101} & \rotatebox{72}{Average}\\ 
			\hline
			(2,1) & 70.40 & 93.70 & 92.30 & 70.10 & 83.20 & 84.30 & 30.20 & 70.40 & 54.30 & 72.30 & 76.30 & 72.50 \\
			(2,2) & 70.43 & 94.27 & 92.62 & 69.40 & 82.87 & 86.08 & 30.25 & 70.17 & 54.77 & 74.19 & 76.37 & 72.86 \\
			(4,1) & 70.44 & 94.12 & \textbf{92.88} & 69.67 & 83.81 & 86.28 & 30.71 & 70.49 & 55.18 & 75.41 & \textbf{76.91} & 73.26 \\
			(4,2) & 70.34 & 94.11 & 92.78 & 69.78 & 83.88 & 86.28 & 30.37 & 70.45 & 54.85 & 74.40 & 76.50 & 73.07 \\
			(4,4) & 70.40 & 94.28 & 92.73 & 69.30 & 83.56 & 86.17 & 30.23 & 70.36 & 54.83 & 73.80 & 76.08 & 72.89 \\
			(8,1) & \textbf{70.45} & \textbf{94.29} & 92.64 & \textbf{70.15} & \textbf{84.44} & \textbf{86.35} & 31.01 & 70.55 & 54.93 & 74.68 & 76.56 & 73.28 \\
			(8,2) & 70.40 & 94.21 & 92.83 & 70.02 & 84.40 & 86.34 & \textbf{31.12} & \textbf{70.66} & \textbf{55.24} & \textbf{75.81} & 76.79 & \textbf{73.44} \\
			(8,4) & 70.33 & 94.08 & 92.55 & 70.08 & 84.42 & 86.34 & 30.95 & 70.57 & 54.67 & 75.37 & 76.32 & 73.24 \\
			(8,8) & 70.35 & 94.21 & 92.83 & 69.52 & 84.11 & 86.21 & 30.43 & 70.37 & 54.83 & 73.13 & 75.95 & 72.90 \\
			\hline
		\end{tabular}
	}
\end{table*}

\textbf{Hyperparameter analysis.} 
As mentioned above, to ensure consistency with the default rank in CLIP-LoRA, Block-LoRA$(2,2)$ is as the default configuration in our experiments, where  the rank $r$ and the number of blocks $n$are both $2$ by default. 
To further investigate the impacts of these hyperparameters on downstream performance, we evaluates the performance of different rank $r$ and number of blocks $n$ in few-shot image classification task across the above 11 datasets.
The results are summarized in Table \ref{tab:ch5_rn}, which contains the classification accuracy under the 1-shot setting. 
Notably, Block-LoRA$(r,n)$ is equivalent to the vanilla LoRA when $n=1$. All experimental settings are consistent with those in the few-shot classification task, and results are averaged over three runs. From Table \ref{tab:ch5_rn}, we observe that: 
First, increasing the rank slightly improves accuracy at most cases. For example, the average accuracy increases from 72.86\% for Block-LoRA$(2,2)$ to 73.07\% for Block-LoRA$(4,2)$ and 73.44\% for Block-LoRA$(8,2)$.  
However, a higher rank also leads to an increase in the number of trainable parameters and computational complexity.
Secondly, using the matrix Block sharing operation and choosing the appropriate number of blocks $n$ will improve the accuracy compared to vanilla LoRA, while maintaining efficiency.

\begin{table}[t] \scriptsize
	\caption{Computing overhead comparison}
	\centering
	\setstretch{2}
	\label{tab:ch5-time}
	\setlength{\tabcolsep}{1.8mm}{
		\begin{tabular}{l | c|c|c|c}
			\hline
			& Parameter & Complexity & Time & Training parameter \\
			\hline
			LoRA-CLIP & $2rd$ & $rd^2$ & 8.7h & 184K \\
			Block-LoRA & $(1+\frac{1}{n})rd$ & $(\frac{1}{n}+\frac{1}{d})rd^2$ & 6.0h & 138K \\
			\hline 
			Proportion & $\frac{1+\frac{1}{n}}{2} $ & $\frac{1}{n}+\frac{1}{d} \approx \frac{1}{n} $ & 69\% &  75\%  \\
			\hline
		\end{tabular}
		}
\end{table}

\begin{figure}[t]
	\centering
	\includegraphics[width=0.8\columnwidth]{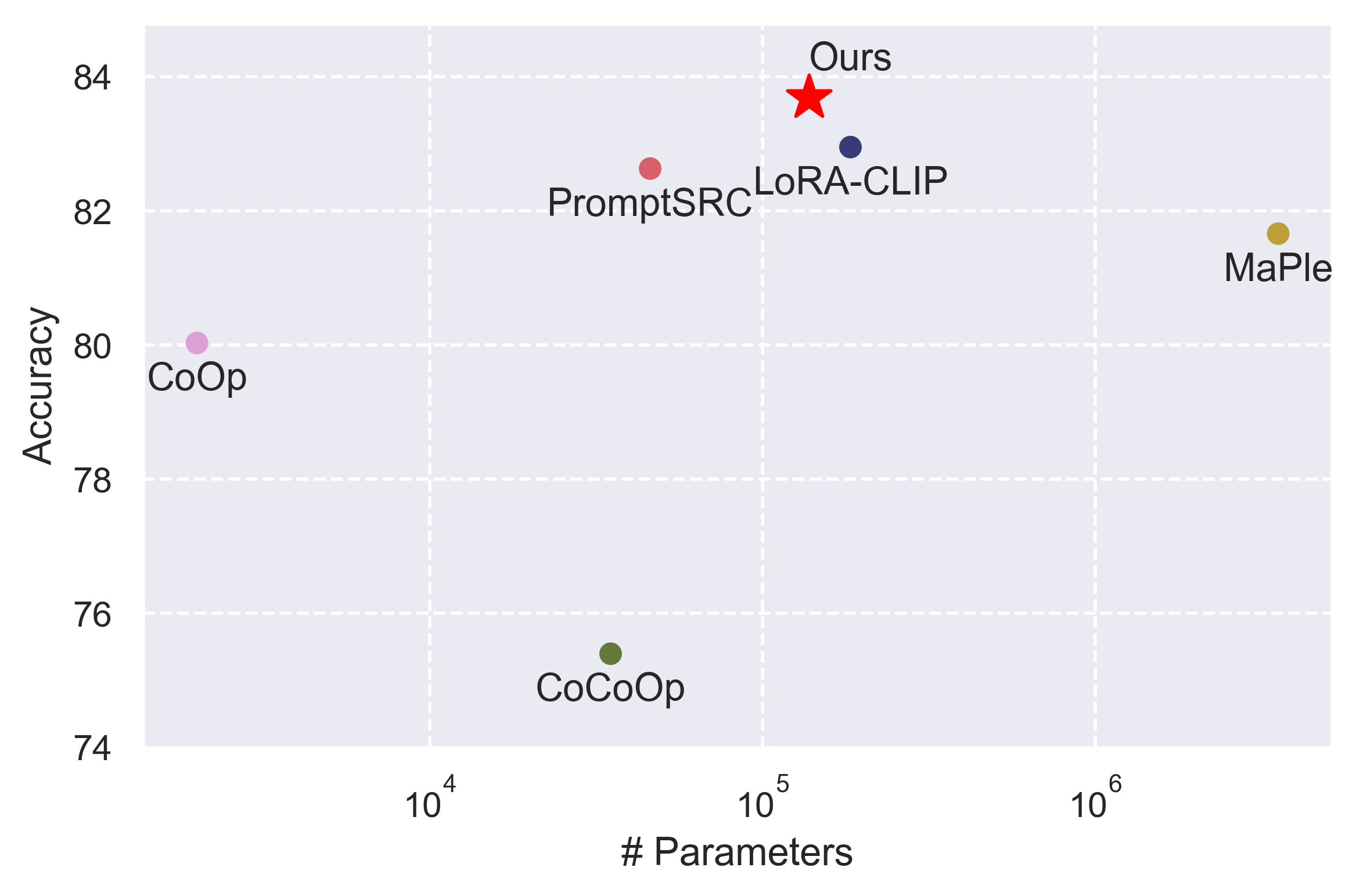}
	\caption{Comparisons of the actual training parameters counts.}
	\label{fig:ch5_parameters}
    \vspace{-0.4cm}
\end{figure}

\noindent\textbf{Compute cost analysis.} 
As discussed above, Block-LoRA reduces trainable parameters and computational complexity compared to vanilla LoRA. To further quantify these differences,  Table \ref{tab:ch5-time} summarizes a comprehensive comparisons of  theoretical parameters count (Parameter for short) and computational complexity (Complexity for short).
Table \ref{tab:ch5-time} also included the actual training time and the actual number of training parameters for Block-LoRA$(2,2)$ under the 16-shot ImageNet setting. 
From Table \ref{tab:ch5-time}, we derive the following insights:
First, the theoretical parameters of  Block-LoRA $(r, n)$ is $\frac{1+\frac{1}{n}}{2}  \in (\,\frac{1}{2} , \frac{3}{4}\,] $ of that in CLIP-LoRA, leading to at least a 25\% reduction.
The number of actual training parameters of Block-LoRA$(2,2)$ used primarily is 184K, exactly $75\%$ of the number of parameters of CLIP-LoRA with $r=2$. Due to the low parameter count, lock-LoRA is device-friendly, running comfortably on a single 24GB GPU even for large-scale ImageNet experiments.
Increasing the number of blocks $n$ leads to a reduction in rank per block $\frac{r}{n}$, further lowering the number of parameters.
Second, for computational complexity, Block-LoRA$(r, n)$ is $(\frac{1}{n}+\frac{1}{d}) \in ( \,\frac{1}{r} , \frac{1}{2}\,) $ of that in CLIP-LoRA, due to $r \le n \ll d$, implying at least a 50\% reduction in computational cost.
Furthermore, Block-LoRA accelerates training, reducing training time to 69\% of CLIP-LoRA on average.
As $n$ increases, the matrix dimensions in computations decrease, which leads to further efficiency gains, which still keep a strong generalization ability, as shown in Table \ref{tab:ch5_rn}.

To visually demonstrate the parameter efficiency of Block-LoRA, Fig \ref{fig:ch5_parameters} presents a scatter plots comparing the number of parameters and the average accuracy on 11 datasets for other methods. First, Block-LoRA achieves the best average accuracy while keeping the number of parameters at a low level. Compared to CLIP-LoRA, Block-LoRA offers better  performance with significantly fewer parameters. 
Second, PromptSRC strikes a reasonable balance between accuracy and parameter count, but its complex training module results in a longer training time (about 15 hours on ImageNet), which is much higher than Block-LoRA. MaPLe incurs a massive parameter overhead due to the use of multi-modal prompts at each network layer. CoOP and CoCoOP, despite their low parameter counts, underperform on downstream tasks.
In summary, Block-LoRA achieves SOTA accuracy while maintaining low parameter count and computational complexity, making it a highly efficient and device-friendly solution for few-shot classification. 

\section{Conclusion}
The recent Vision-Language Foundation Models (VLMs) based  few-shot learning have achieved promising results. However, these models often  suffer from excessive training parameters and high computational costs.  To address these challenges, we propose a novel Block matrix-based low-rank adaptation framework, Block-LoRA, for fine-tuning VLMs on downstream few-shot tasks. Block-LoRA partitions the original low-rank decomposition matrix of LoRA into multiple sub-matrices while sharing all down-projection sub-matrices. This structure not only reduces the number of trainable parameters but also simplifies certain complex matrix multiplications into more efficient matrix additions, significantly lowering the computational cost of fine-tuning. Extensive experiments demonstrate that Block-LoRA achieves competitive performance compared to SOTA CLIP-based few-shot methods, while maintaining a low training parameters count and reduced computational overhead. The effectiveness of Block-LoRA on other visual or textual tasks deserves further exploration.

\bibliographystyle{named}
\bibliography{ijcai25.bib}

\cleardoublepage

\appendix
\begin{center}\LARGE
\textbf{Appendix}    
\end{center}

\section{Proof of Lemma 1}
We first proof the generalization upper bound of vanilla LoRA \cite{hu2022lora}. Following previous works \cite{xu2017information,zhu2024asymmetry}, we have the bound as follows:

\noindent \textbf{Theorem 1.} The LoRA-based fine-tuning algorithm has an adaptation matrices set as \(\Delta \boldsymbol{W} =\{\Delta \mathbf{W}_{l}\}_{l = 1}^{L}\), trained by dataset $S$.  Assume that the loss \(\ell^{\boldsymbol{W}}(\Delta \boldsymbol{W},Z)\) is \(\sigma\)-sub-Gaussian under \((\Delta \boldsymbol{W},Z)
\sim P_{\Delta \boldsymbol{W} |\boldsymbol{W} }\times\mu\). Then,

\begin{equation}\label{eq:a1}
|error(\mu, \text{LoRA})| \leq \sqrt{\frac{2 \sigma^{2}}{\#S} I\left(\Delta \boldsymbol{W} ; S \ | \text{LoRA}, \boldsymbol{W} \right)}. 
\end{equation}
where $I(\ )$ denotes mutual information. And we have:

\begin{equation*}
 I\left(\Delta \boldsymbol{W} ; S \ | \text{LoRA}, \boldsymbol{W} \right)  =  
 I\left( \{\mathbf{A}_{l}\mathbf{B}_{l}\}_{l\in I} ; S \ | \text{LoRA}, \boldsymbol{W} \right)  
\end{equation*}
Further considering definition of mutual information: 
\begin{equation*}
 I(X;Y) = H(X) - H(X|Y) \le  H(X)
\end{equation*}
We further have:
\begin{equation*}
\begin{aligned}
 I\left( \{\mathbf{A}_{l}\mathbf{B}_{l}\}_{l\in I} ; S \ | \text{LoRA}, \boldsymbol{W} \right)  &\le H  ( \{\mathbf{A}_{l}\mathbf{B}_{l}\}_{l\in I} ) \\
 &\le  qr\sum_{l\in\mathcal{I}}\left(k_{(l)} + d_{(l)} \right) \ln 2
\end{aligned}
\end{equation*}
Substitute the above inequation into Eq (\ref{eq:a1}), we obtain the the bounds for LoRA algorithm:
\begin{align*}
\left|error\left(\mu,\text{LoRA}\right)\right|&\leq\sqrt{\frac{2rq\sigma^{2}\ln 2}{\#S}\sum_{l\in\mathcal{I}}\left(k_{(l)} + d_{(l)} \right)}\\
\end{align*}

Next, we proof the generalization upper bound of the proposed Block-LoRA in this paper. Similar with LoRA, we have:

\noindent \textbf{Theorem 2.} The Block-LoRA-based fine-tuning algorithm (Block for short) has an adaptation matrices set as \(\Delta \boldsymbol{W} =\{\Delta \mathbf{W}_{l}\}_{l = 1}^{L}\), trained by dataset $S$.  Assume that the loss \(\ell^{\boldsymbol{W}}(\Delta \boldsymbol{W},Z)\) is \(\sigma\)-sub-Gaussian under \((\Delta \boldsymbol{W},Z)
\sim P_{\Delta \boldsymbol{W} |\boldsymbol{W} }\times\mu\). Then,

\begin{equation}\label{eq:a2}
|error(\mu, \text{Block})| \leq \sqrt{\frac{2 \sigma^{2}}{\#S} I\left(\Delta \boldsymbol{W} ; S \ | \text{Block}, \boldsymbol{W} \right)}. 
\end{equation}
We further have:
\begin{equation*}
\begin{aligned}
 &I\left(\Delta \boldsymbol{W} ; S \ | \text{Block}, \boldsymbol{W} \right)  \\
 =  
 &I\left( \{\mathbf{A_s}_l \sum_{i=1}^{n}   \mathbf{B_i}_l \}_{l\in I} ; S \ | \text{Block}, \boldsymbol{W} \right) \\
 \le &H( \{\mathbf{A_s}_l \sum_{i=1}^{n}   \mathbf{B_i}_l \}_{l\in I} ) \\
\le &qr\sum_{l\in\mathcal{I}} (\frac{k_{(l)}}{n} + d_{(l)})\ln 2
 \end{aligned}
\end{equation*}
Substitute the above inequation into Eq (\ref{eq:a2}), we obtain the the bounds for Block-LoRA algorithm:
\begin{align*}
\left|error \left(\mu,\text{Block-LoRA}\right)\right|&\leq\sqrt{\frac{2rq\sigma^{2}\ln 2}{\#S}\sum_{l\in\mathcal{I}}\left(\frac{k_{(l)}}{n} + d_{(l)} \right)}
\end{align*}

\begin{table*}[t] \small 
	\centering
	\setstretch{1.5}
	\caption{ Statistics of 11 datasets for few-shot learning and 4 target datasets of domain generalization.}
	\begin{tabular}{c|c|c|c}
		\hline Datasets &  \# Classes & Train / Val / Test & Description  \\
		\hline ImageNet \cite{russakovsky2015imagenet} & 1000 & 1.28M / - /50000 & Recognition of generic objects \\
		Caltech101 \cite{fei2004learning} & 100 & 4128 / 1649 / 2465 & Recognition of generic objects\\
		OxfordPets \cite{parkhi2012cats} & 37 & 2944 / 736 / 3669 &  Fine-grained classification of pets  \\
		StanfordCars  \cite{krause20133d}& 196 & 6509 / 1635 / 8041 &  Fine-grained classification of cars \\
		Flowers102 \cite{nilsback2008automated} & 102 & 4093 / 1633 / 2463 & Fine-grained classification of flowers\\
		Food101 \cite{bossard2014food} & 101 & 50500 / 20200 / 30300 & Fine-grained classification of foods \\
		FGVCAircraft \cite{maji2013fine} & 100 & 3334 / 3333 / 3333 & Fine-grained classification of aircrafts \\
		SUN397 \cite{xiao2010sun} & 397 & 15880 / 3970 / 19850 &  Scene classification\\
		DTD \cite{cimpoi2014describing} & 47 & 2820 / 1128 / 1692 & Texture classification \\
		EuroSAT \cite{helber2019eurosat}& 10 & 13500 / 5400 / 8100 &   Land cover classification with satellite images \\
		UCF101 \cite{soomro2012ucf101}& 101 & 7639 / 1898 / 3783 &  Action recognition \\
		\hline ImageNet-V2  \cite{recht2019imagenet}& 1000 & -/ - / 10000 & New test data for ImageNet \\
		ImageNet-Sketch \cite{wang2019learning}& 1000 & -/ - / 50889 & Sketch-style images of ImageNet classes\\
		ImageNet-A \cite{hendrycks2021natural}& 200 & -/ - / 7500 & Natural adversarial examples of  ImageNet classes\\
		ImageNet-R \cite{hendrycks2021many} & 200 & -/ - / 30000 &  Renditions of 200 ImageNet classes\\
		\hline
	\end{tabular}
	\label{tab:ch5_datasets}
\end{table*}

\section{Details of Datasets}

\subsection{Few-shot learning dataset}
This paper employs 11 publicly available image datasets for few-shot learning tasks, consistent with previous works \cite{zhou2022learning,zanella2024low}. These datasets include:

\textbf{Datasets for Generic  Object Recognition Tasks}: The ImageNet \cite{russakovsky2015imagenet} and Caltech101 \cite{fei2004learning} datasets. ImageNet is a large-scale dataset widely used in the field of computer vision for general object recognition, containing a vast number of images across 1000 object categories. 
Caltech101, also serves as a benchmark for general object recognition tasks, with 101 object categories.

\textbf{Datasets for Fine-Grained Image Classification Tasks}: The OxfordPets \cite{parkhi2012cats}, StanfordCars \cite{krause20133d}, Flowers102 \cite{nilsback2008automated}, Food101 \cite{bossard2014food}, and FGVCAircraft \cite{maji2013fine} datasets. OxfordPets focuses on classifying pet images, which requires fine-grained discrimination between different breeds. StanfordCars is dedicated to classifying car models, where the differences between classes are often subtle. Flowers102 is designed for flower classification, and the fine-grained nature of flower species demands high precision classification. Food101 is used for classifying different types of food images, and FGVCAircraft is specialized in fine-grained classification of aircraft images.

 \textbf{Dataset for Scene Recognition Task}: SUN397 \cite{xiao2010sun} dataset is specifically designed to assist in scene recognition tasks, covering a wide range of natural and man-made scenes, enabling models to learn the characteristics of different scenes.

 \textbf{Dataset for Action Recognition Task}: The UCF101 \cite{soomro2012ucf101} dataset contains a large number of video sequences related to 101 different human actions.

 \textbf{Dataset for Texture Classification Task}: The DTD \cite{cimpoi2014describing} dataset is focused on texture classification, with a collection of images that represent various textures, allowing models to learn to distinguish between different texture patterns.

\textbf{Dataset for Satellite-based Land Use and Cover Classification Task}: The EuroSAT \cite{helber2019eurosat} dataset consists of satellite images for land use and cover classification, helping to analyze and classify different types of land use from satellite-based imagery.

\subsection{Domain generalization dataset}
For the domain generalization task, this paper uses 4 publicly available domain generalization datasets, all of which are variants of the ImageNet dataset. These include:

 \textbf{ImageNetV2} \cite{recht2019imagenet} dataset is a newly re-collected test data, ensuring its difference from previous datasets. It aims to provide a more up-to-date and independent test set for evaluating the generalization ability of models.
 
 \textbf{ImageNet-Sketch} \cite{wang2019learning} dataset is composed of black-and-white sketch images collected from Google Image Search. After collection, manual screening is carried out to ensure the correctness of the data. Sketch images present a different visual modality compared to the original ImageNet color images, challenging models to generalize across different visual representations.
 
\textbf{ImageNet-A(dversarial)}\cite{hendrycks2021natural} dataset contains natural adversarial examples collected from the real world. These natural adversarial examples often cause machine learning models to make incorrect predictions. By using this dataset, we can test the robustness of models against such real-world adversarial inputs.

 \textbf{ImageNet-R(endition)} \cite{hendrycks2021many} dataset consists of category-variant images from different art styles such as cartoons, doodles, origami, and tattoos. It provides a means to assess how well models can generalize across different artistic renditions of the same object categories.

The detailed statistical data of all the above datasets are presented in Table \ref{tab:ch5_datasets}.

\section{Ablation Study}
As shown in Eq (\ref{eq:9}), the proposed Block-LoRA introduces a learnable shared down-projection matrix $\mathbf{A_s}$. 
Recent studies on LoRA consistently suggest that down-projection matrix  matrices $\mathbf{A}$ are generally less important than up-projection matrix  matrices $\mathbf{B}$  \cite{zhu2024asymmetry,zhang2023lora}. To investigate this further, we evaluate a variant of Block-LoRA, denoted as w/o A, where $\mathbf{A}$ is randomly initialized and remains fixed during training (i.e., it is not updated). We conduct experiments on four datasets, testing performance across 1-shot to 16-shot settings while keeping all other experimental configurations consistent with before. The results are summarized in Table \ref{tab:ch5_noA}, leading to the following observations:
\begin{enumerate}
    \item First, Block-LoRA, which employs a learnable shared down-projection matrix $\mathbf{A_s}$, outperforms its w/o A variant in most cases. This indicates that completely freezing $\mathbf{A_s}$ may negatively impacts performance for few-shot learning tasks.

    \item Second, the performance gap is more pronounced in low-data scenarios (e.g., 1-shot). However, as the number of training samples increases (e.g., 16-shot), the gap narrows significantly, suggesting that increased data availability compensates for the absence of a learnable $\mathbf{A_s}$. To the best of our knowledge, we are the first to find this phenomenon that the performance gap is related to the data scale.

    \item Third, as the number of trainable parameters in the model increases (e.g., by increasing the matrix rank $r$ ), the performance gap between Block-LoRA and w/o A further decreases. This implies that a larger up-projection matrix $\mathbf{B}$ can partially offset the negative effects of the absence of a learnable $\mathbf{A_s}$.
\end{enumerate}

In summary, completely freezing the down-projection matrix $\mathbf{A_s}$ may degrade downstream task performance. Thus, Block-LoRA employs a learnable $\mathbf{A_s} \in  \mathbb{R}^{k  \times \frac{r}{n} }$  defaultly.  Moreover, compared to vanilla LoRA, which uses a larger down-projection matrix $\mathbf{A}  \in  \mathbb{R}^{k  \times r } $, Block-LoRA achieves highly competitive performance with fewer parameters.

\begin{table*}[!ht] \footnotesize
	\centering
	\caption{Effects of the absence of the learnable $\mathbf{A_s}$ in Block-LoRA.}
	\setstretch{1.3}
	\label{tab:ch5_noA}
	\setlength{\tabcolsep}{1.2mm}{
		\begin{tabular}{c|c|cc|cc|cc|cc}
			\hline
			\multicolumn{1}{l}{} & \multicolumn{1}{l}{Dataset} & \multicolumn{2}{c}{EuroSAT} & \multicolumn{2}{c}{ImageNet} & \multicolumn{2}{c}{SUN397} & \multicolumn{2}{c}{UCF101} \\
			\hline
			& (r,n) & w/o A& Block-LoRA& w/o A & Block-LoRA & w/o A & Block-LoRA& w/o A & Block-LoRA \\
			\hline
			\multirow{9}{*}{1-shot} & (2,1) & 69.45 & 72.30 & 68.65 & 70.40 & 68.16 & 70.40 & 71.34 & 76.30 \\
			& (2,2) & 69.17 & 74.19 & 68.62 & 70.43 & 67.71 & 70.17 & 70.99 & 76.37 \\
			& (4,1) & 71.74 & 76.02 & 69.01 & 70.44 & 68.58 & 70.49 & 72.38 & 76.91 \\
			& (4,2) & 71.50 & 74.40 & 69.02 & 70.34 & 68.62 & 70.45 & 72.52 & 76.50 \\
			& (4,4) & 70.87 & 73.80 & 68.94 & 70.40 & 68.28 & 70.36 & 72.37 & 76.08 \\
			& (8,1) & 72.45 & 74.68 & 69.51 & 70.45 & 69.37 & 70.55 & 73.58 & 76.56 \\
			& (8,2) & 73.17 & 75.81 & 69.40 & 70.40 & 69.23 & 70.66 & 73.42 & 76.79 \\
			& (8,4) & 72.76 & 75.37 & 69.36 & 70.33 & 69.18 & 70.57 & 73.33 & 76.32 \\
			& (8,8) & 72.48 & 73.13 & 69.21 & 70.35 & 68.84 & 70.37 & 73.24 & 75.95 \\
			\hline
			\multirow{9}{*}{2-shot} & (2,1) & 80.39 & 82.70 & 69.45 & 70.80 & 68.55 & 71.30 & 75.26 & 80.00 \\
			& (2,2) & 80.67 & 83.94 & 69.31 & 70.79 & 68.27 & 71.71 & 74.87 & 79.76 \\
			& (4,1) & 81.29 & 83.21 & 69.82 & 70.78 & 69.16 & 71.97 & 76.25 & 80.14 \\
			& (4,2) & 81.23 & 83.89 & 69.75 & 70.84 & 69.09 & 72.02 & 76.44 & 79.98 \\
			& (4,4) & 81.77 & 83.63 & 69.61 & 70.73 & 68.88 & 71.78 & 76.24 & 79.95 \\
			& (8,1) & 82.45 & 83.99 & 70.19 & 70.92 & 69.93 & 72.28 & 77.85 & 79.92 \\
			& (8,2) & 81.58 & 82.73 & 70.07 & 70.83 & 69.73 & 72.06 & 77.80 & 79.95 \\
			& (8,4) & 81.74 & 83.77 & 70.04 & 70.88 & 69.78 & 72.08 & 77.55 & 79.85 \\
			& (8,8) & 82.16 & 82.73 & 69.88 & 70.73 & 69.43 & 71.80 & 77.35 & 79.75 \\
			\hline
			\multirow{9}{*}{4-shot} & (2,1) & 83.07 & 84.90 & 70.14 & 71.40 & 69.95 & 72.80 & 78.55 & 81.10 \\
			& (2,2) & 84.77 & 88.08 & 70.08 & 71.53 & 69.69 & 73.43 & 77.75 & 82.12 \\
			& (4,1) & 84.79 & 86.82 & 70.57 & 71.63 & 70.92 & 73.75 & 79.97 & 82.25 \\
			& (4,2) & 84.03 & 87.34 & 70.45 & 71.53 & 70.75 & 73.53 & 79.74 & 81.99 \\
			& (4,4) & 85.63 & 88.06 & 70.36 & 71.43 & 70.36 & 73.39 & 79.18 & 82.10 \\
			& (8,1) & 85.61 & 88.34 & 70.92 & 71.76 & 71.89 & 73.95 & 81.09 & 82.18 \\
			& (8,2) & 85.28 & 87.26 & 70.90 & 71.70 & 71.70 & 73.73 & 81.02 & 82.16 \\
			& (8,4) & 84.64 & 87.44 & 70.76 & 71.54 & 71.44 & 73.59 & 80.74 & 81.87 \\
			& (8,8) & 86.13 & 87.98 & 70.64 & 71.49 & 71.03 & 73.34 & 80.22 & 82.10 \\
			\hline
			\multirow{9}{*}{8-shot} & (2,1) & 88.69 & 89.70 & 70.90 & 72.30 & 71.74 & 74.70 & 81.62 & 84.10 \\
			& (2,2) & 88.54 & 90.56 & 70.81 & 72.39 & 71.25 & 75.06 & 81.19 & 84.07 \\
			& (4,1) & 89.30 & 90.89 & 71.43 & 72.39 & 72.57 & 75.28 & 83.14 & 84.72 \\
			& (4,2) & 89.26 & 90.67 & 71.20 & 72.48 & 72.58 & 75.15 & 82.58 & 84.44 \\
			& (4,4) & 88.95 & 90.78 & 71.14 & 72.32 & 72.02 & 75.02 & 82.10 & 84.10 \\
			& (8,1) & 89.93 & 91.05 & 71.83 & 72.63 & 73.45 & 75.43 & 84.16 & 85.06 \\
			& (8,2) & 89.78 & 91.08 & 71.77 & 72.62 & 73.36 & 75.29 & 83.68 & 84.57 \\
			& (8,4) & 89.50 & 91.05 & 71.52 & 72.50 & 73.25 & 75.16 & 83.36 & 84.43 \\
			& (8,8) & 89.30 & 91.08 & 71.43 & 72.33 & 72.66 & 75.02 & 82.84 & 84.10 \\
			\hline
			\multirow{9}{*}{16-shot} & (2,1) & 92.88 & 92.10 & 71.74 & 73.60 & 73.49 & 76.10 & 84.83 & 86.70 \\
			& (2,2) & 92.65 & 93.32 & 71.59 & 73.47 & 73.02 & 76.77 & 84.40 & 87.09 \\
			& (4,1) & 92.72 & 93.40 & 72.27 & 73.63 & 74.47 & 76.97 & 85.82 & 86.89 \\
			& (4,2) & 93.09 & 93.65 & 72.10 & 73.53 & 74.26 & 76.98 & 85.43 & 87.16 \\
			& (4,4) & 92.83 & 93.34 & 71.89 & 73.32 & 73.73 & 76.73 & 85.00 & 86.96 \\
			& (8,1) & 93.35 & 93.76 & 72.85 & 73.80 & 75.51 & 77.09 & 86.63 & 87.41 \\
			& (8,2) & 92.85 & 93.32 & 72.65 & 73.71 & 75.15 & 76.94 & 86.08 & 87.03 \\
			& (8,4) & 93.20 & 93.67 & 72.41 & 73.55 & 74.99 & 76.99 & 86.13 & 87.21 \\
			& (8,8) & 93.01 & 93.43 & 72.18 & 73.36 & 74.42 & 76.72 & 85.67 & 86.90 \\
			\hline
		\end{tabular}
	}
\end{table*}

\end{document}